\documentclass{article}

\usepackage{microtype}
\usepackage{graphicx}
\usepackage{booktabs} 
\usepackage{xcolor}

\usepackage{stfloats}
\usepackage{amsmath}
\usepackage{float}
\usepackage[table]{xcolor}   

\usepackage[breaklinks=true]{hyperref}
\usepackage{bclogo}

\expandafter\def\expandafter\UrlBreaks\expandafter{\UrlBreaks
    \do\a\do\b\do\c\do\d\do\e\do\f\do\g\do\h\do\i\do\j%
    \do\k\do\l\do\m\do\n\do\o\do\p\do\q\do\r\do\s\do\t%
    \do\u\do\v\do\w\do\x\do\y\do\z\do\A\do\B\do\C\do\D%
    \do\E\do\F\do\G\do\H\do\I\do\J\do\K\do\L\do\M\do\N%
    \do\O\do\P\do\Q\do\R\do\S\do\T\do\U\do\V\do\W\do\X%
    \do\Y\do\Z\do\/\do-}

\usepackage[accepted]{mlsys2025}

\usepackage{xspace}
\usepackage{amsfonts}
\usepackage{amsmath}
\usepackage[capitalize, noabbrev]{cleveref}
\usepackage[user, hyperref,savepos]{zref}
\usepackage{subcaption}
\usepackage{multirow}
\usepackage{graphicx} 

\newcommand{\method}{\texttt{ScaleSearch}\xspace}
\newcommand{\methodattention}{\texttt{ScaleSearchAttention}\xspace}
\newcommand{\methodbold}{\textbf{\texttt{ScaleSearch}}\xspace}

\mlsystitlerunning{Search Your Block Floating Point Scales!}

\usepackage{listings}

\begin{document}

\twocolumn[
\mlsystitle{\bcloupe{} Search Your Block Floating Point Scales!}



\mlsyssetsymbol{equal}{*}

\begin{mlsysauthorlist}
\mlsysauthor{Tanmaey Gupta}{cu,to}
\mlsysauthor{Hayden Prairie}{ucsd,to}
\mlsysauthor{Xiaoxia Wu}{to}
\mlsysauthor{Reyna Abhyankar}{to}
\mlsysauthor{Qingyang Wu}{to}
\mlsysauthor{Austin Silveria}{to}
\mlsysauthor{Pragaash Ponnusamy}{to}
\mlsysauthor{Jue Wang}{to}
\mlsysauthor{Ben Athiwaratkun}{to}
\mlsysauthor{Leon Song}{to}
\mlsysauthor{Tri Dao}{princeton,to}
\mlsysauthor{Daniel Y. Fu}{ucsd,to}
\mlsysauthor{Chris De Sa}{cu,to}
\end{mlsysauthorlist}

\mlsysaffiliation{cu}{Cornell University, Ithaca, NY, USA}
\mlsysaffiliation{ucsd}{University of California San Diego, La Jolla, CA, USA}
\mlsysaffiliation{princeton}{Princeton University, Princeton, NJ, USA}
\mlsysaffiliation{to}{Together AI, USA}

\mlsyscorrespondingauthor{Tanmaey Gupta}{tanmaeygupta99@gmail.com}

\mlsyskeywords{Machine Learning, MLSys}

\vskip 0.3in

\begin{abstract}
Quantization has emerged as a standard technique for accelerating inference for generative models by enabling faster low-precision computations and reduced memory transfers. Recently, GPU accelerators have added first-class support for microscaling Block Floating Point (BFP) formats.  Standard BFP algorithms use a fixed scale based on the maximum magnitude of the block. We observe that this scale choice can be suboptimal with respect to quantization errors. In this work, we propose \method, an alternative strategy for selecting these scale factors: using a fine-grained search leveraging the mantissa bits in microscaling formats to minimize the quantization error for the given distribution. \method can be integrated with existing quantization methods such as Post Training Quantization and low-precision attention, and is shown to improve their performance. Additionally, we introduce \methodattention, an accelerated NVFP4-based attention algorithm, which uses \method and adapted prior techniques to ensure near-0 performance loss for causal language modeling. Experiments show that \method reduces quantization error by 27\% for NVFP4 and improves language model PTQ by up to 15 points for MATH500 (Qwen3-8B), while \methodattention improves Wikitext-2 PPL by upto 0.77 points for Llama 3.1 70B. The proposed methods closely match baseline performance while providing quantization accuracy improvements.
\end{abstract}
]



\printAffiliationsAndNotice{}  
 \section{Introduction}
Quantization plays a key role in optimizing inference efficiency for generative models \cite{quant_survey} — though maintaining accuracy at ultra-low bit-widths is challenging to achieve. Block Floating Point (BFP) formats have recently gained prominence over fixed-point and floating-point formats, offering a balanced trade-off between dynamic range, memory efficiency, and computational throughput \cite{bfp1, msfp, fast}. Recent advancements have improved the algorithmic accuracy and hardware support for low-bitwidth BFP formats. In particular, NVIDIA’s Blackwell architecture \cite{nvfp4} introduces fast FP4 arithmetic through NVFP4 and MXFP4 \cite{mxfp4} microscaling formats, enabling 4-bit matrix multiplications directly on Tensor Cores, achieving $2 \times$ higher throughput compared to FP8 compute on B200 GPUs and $3 \times$ higher throughput on the B300. A critical question is how to best use these formats to optimize machine learning models. Existing BFP implementations in both research and production frameworks (e.g., TensorRT, vLLM) \cite{tensorrt, vLLM, shared_exponents, msfp, bfp1, quant_survey, svdquant} predominantly rely on a reasonable baseline, where the block scale is determined by the maximum absolute value in each block, which maps the values to the range representable by the low-precision format. We observe that this approach can be suboptimal, and alternative scale choices can lower quantization errors.

\begin{figure*}[t]
\includegraphics[width=0.9\textwidth]{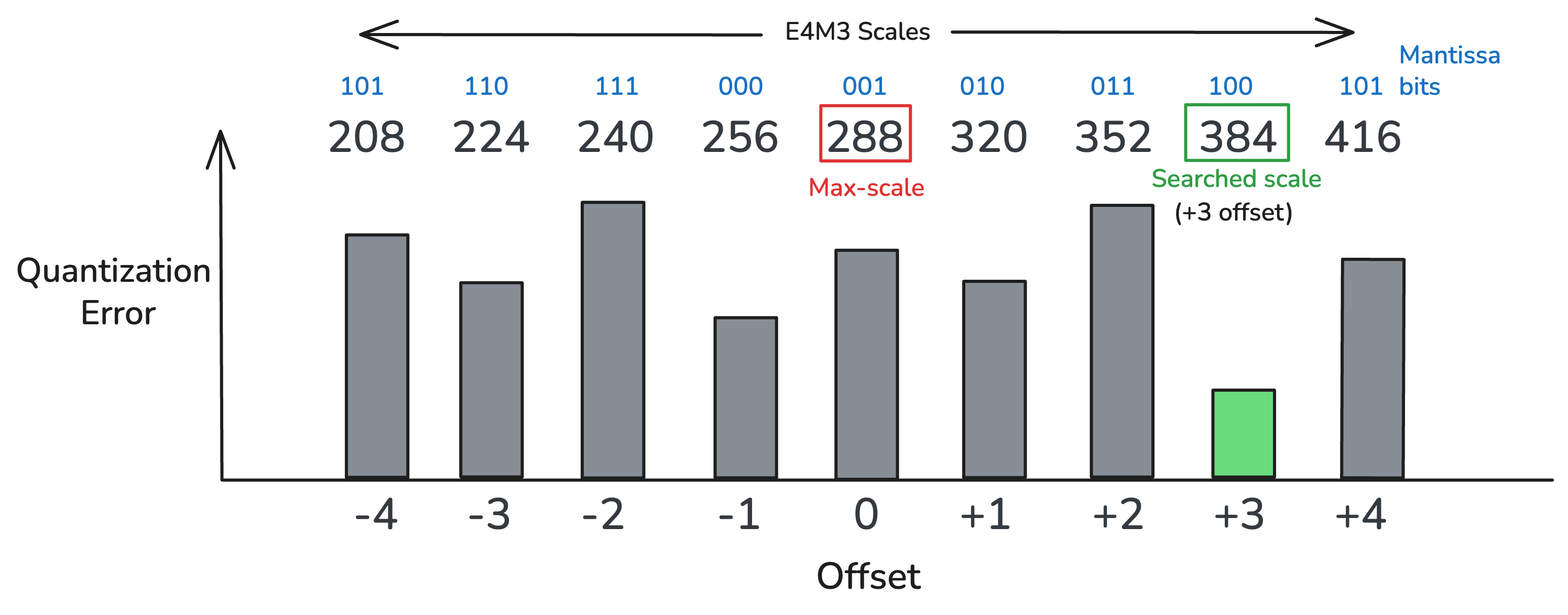}
\caption{\method searches for a block scale which gives the minimum quantization error.}
\label{fig:scalesearch}
\end{figure*}

In this paper, we propose an alternate approach, \method (shown \cref{fig:scalesearch}). Recent microscaling BFP formats like NVFP4 introduce mantissa bits in block scales (E4M3 Floating Point), which can be employed for fine-grained tuning of the scale. Building upon this insight, \method searches for the optimal block scale to minimize quantization error for a given distribution. \method is architecture-agnostic and can be integrated into various quantization pipelines. We demonstrate its benefits across Post-Training Quantization (PTQ) of weights and activations, as well as in low-precision attention computation.

Further, we extend FP4 optimizations to the attention mechanism and KV cache—components that dominate inference-time memory and compute costs due to their quadratic complexity \cite{fa1}. While recent works have explored FP4 quantization for weights and activations in PTQ and QAT \cite{llmcompressor, trt_opt}, and even full FP4-based training \cite{fp4train1,fp4train2,fp4train3} with negligible accuracy loss \cite{blackwell_infer}, FP4-native attention and KV cache compression remain underexplored. 

To this extent, we propose \methodattention, an extension to \method that enables NVFP4 quantization of the KV cache and attention mechanism for causal language modeling. We use \method to select scale values for queries, keys, and values in attention, as well as the partial attention matrix in FlashAttention \cite{fa1}. \methodattention enables the use of fast NVFP4 Tensor Core operations without dequantization for $QK^T$ and $PV$ attention matrix multiplications. To achieve near-0 model accuracy degradations, \methodattention further employs Incoherence Processing \cite{quip, quipsharp} and matrix decomposition to reduce outliers and the average magnitude of Q and K tensors, which in turn reduces quantization error. We also use attention-sink-aware \cite{spa5} mixed precision caching, where the first few tokens and the most recent tokens are stored in full precision, and are quantized to NVFP4 once enough tokens have been generated.

We evaluate \method across three common FP4 quantization settings: weights PTQ, low-precision attention for diffusion model generation and causal language modeling. For weights PTQ, \method improves model performance by upto 15 points for MATH500 (Qwen3-8B). For FP4 attention, \method improves upon SageAttention3 \cite{sageattention3} by 14 points for VQA-t performed using the Mochi model. \methodattention reduces PPL from 3.4 to 2.6348 for Llama 3.1 70B and improves the accuracy on GPQA Diamond test benchmark by 5 points for Llama 8.1B Instruct model. The proposed methods closely match baseline
performance while providing quantization accuracy improvements. \method introduces a minimal overhead of 1.74x during FP32 to NVFP4 quantization, and achieves 98.3\% of SageAttention3's attention throughout.

In summary, we make the following contributions in this paper: \begin{enumerate}
    \item We introduce \method, a scale-searching algorithm for Block Floating Point quantization that minimizes block-wise quantization error by exploring neighboring representable scales. Our analysis is primarily based on NVFP4’s E4M3 format where, unlike traditional max-scaling, \method leverages the mantissa resolution of the block scale to select a quantizer that achieves the lowest mean-squared error. We show that \method reduces quantization error by 26\% for synthetic gaussian data, seamlessly integrates into existing PTQ and attention workflows and consistently improves quantized model performance.
    \item We propose \methodattention, a co-optimized FP4 attention mechanism that performs both $QK^T$ and $PV$ matrix multiplications directly in NVFP4 precision on Tensor Cores—without any de-quantization overhead—and stores the KV cache in compact 4.5-bit NVFP4 format. \methodattention employs \method and other adapted quantization techniques, enabling end-to-end FP4 inference while preserving near-zero accuracy degradation.
    \item We conduct extensive experiments on large-scale language models to evaluate the efficacy of \method and \methodattention. Our results demonstrate consistent gains in perplexity and benchmark accuracy over prior FP4 baselines such as SageAttention3 for both language and diffusion models.
\end{enumerate}

\section{Related Work}
\label{sec:related_work}
\textbf{Block scaling quantization for DNN}
Quantized number representation format is a critical design choice that provides a spectrum across dimensions of precision, dynamic range, and hardware support. 
Fixed-point numbers \cite{fixed_point_1,fixed_point2} have fast hardware implementation, but their limited dynamic range is not suitable for representing outliers seen in DNN models. Floating point numbers \cite{fp8_nvidia}, on the other hand, have high computation costs due to the required alignment of mantissa for MAC operations, and narrow representations have low dynamic range. Hence, Block Floating Point (BFP) formats have recently emerged as a compelling middle ground between these two extremes \cite{bfp1, msfp, fast}. The central idea of BFP is that elements within a tensor block often share similar magnitudes and therefore can be represented by a shared exponent and individual integer mantissas. This shared-exponent structure drastically reduces storage overhead and enables the use of integer matrix multiplication pipelines while preserving much of the dynamic range flexibility of floating-point representations. 

Recent advances have led to microscaled BFP formats such as MXFP4, NVFP4, and MXFP8 \cite{mxfp4, nvfp4}, which refine traditional block scaling for modern accelerators. These formats introduce compact per-block scale encodings optimized for Tensor Core operations, enabling efficient low-bit arithmetic without dequantization. Unlike earlier BFP variants that stored only exponent bits, NVFP4 \cite{nvfp4} uniquely incorporates mantissa bits within the scale representation, offering finer granularity and improved dynamic range at minimal overhead. Standard BFP and microscaled formats employ max-based scaling, where the block scale is derived from the maximum absolute value within each block—ensuring representability \cite{msfp, bfp1, quant_survey}.

\textbf{Quantization for Generative Models}
Quantization is a fundamental technique which reduces numerical precision across different model components and targets optimization across dimensions like memory transfer, matrix-multiplication speeds, and end-to-end training time. Broadly, quantization approaches fall into three categories: post-training quantization (PTQ), quantization-aware training (QAT), and fully quantized training. Post-training quantization (PTQ) algorithms like GPTQ \cite{gptq}, AWQ \cite{awq}, and ZeroQuant \cite{zeroquant} quantize model weights post-training, often using second-order error metrics, activation-aware heuristics, or outlier-channel handling to minimize degradation. SmoothQuant \cite{smoothquant} extends PTQ to activations by learning per-channel affine transformations that reduce activation variance. Works like Qserve~\cite{qserve} and llm.int8()\cite{llmint8} demonstrate mixed-precision quantization for different components or outliers. QuIP \cite{quip} advances PTQ by using adaptive rounding and incoherence processing, while QuIP\# \cite{quipsharp} introduces Hadamard transforms and E8-lattice vector quantization.
We adapt the incoherence processing step in our method from these works. In contrast, quantization-aware training (QAT) \cite{qat1,qat2} incorporates quantization simulation into the forward pass during training and which makes low-bit inference robust to quantization loss.   Finally, fully quantized training—which quantizes weights, activations, and gradients—has become feasible due to advances in low-bit numerical formats \cite{fp4train1,fp4train2}. Recent innovations, such as FP4 training using NVFP4 and MXFP4 formats, enable training and inference on 4-bit floating-point hardware.

\textbf{KV Cache Compression} As the KV cache increases linearly with sequence length and batch size, significant efforts have been put into reducing its memory footprint. 
Architecture variants like sparse transformers \cite{spa1,spa2}, low-rank approximations \cite{performers}, and shared KV head methods \cite{mqa,gqa,mhla} inherently reduce the cache size. 
In a parallel direction of using a subset of KV cache, StreamingLLM \cite{spa5} stores only the initial and recent tokens, and methods like Gist Tokens and SnapKV \cite{snapkv,gist} compress the initial prompt into "important tokens" based on learning-based methods. H2O \cite{h20}, RocketKV \cite{rocketkv}, and FastGen \cite{discard} take an alternative approach of dynamic identification of crucial tokens and eviction. Another approach \cite{minicache, kvsharer} shares the KV cache between layers and may additionally store relative vectors.

Previous KV-cache quantization methods face challenges from attention dynamics and outlier channels, making slow compression (e.g., adaptive rounding, lattice codebooks) impractical for high-throughput inference. Token-wise methods like ZipCache \cite{zipcache} and WKVQuant \cite{wkvquant} localize errors, while channel-wise approaches such as KVQuant \cite{kvquant} and KIVI \cite{kivi} better adapt to magnitude variation. Error correction techniques include Gear’s low-rank modeling \cite{gear} and QuaRot’s orthogonal transformations \cite{quarot}. Mixed-precision strategies (e.g., KIVI \cite{kivi}, IntactKV \cite{intactkv}, KVTuner \cite{kvtuner}) statically assign higher precision to critical tokens, but their fixed heuristics and runtime complexity limit scalability. Dynamic approaches (MiKV \cite{mikv}, ZipCache \cite{zipcache}) incur overheads and are incompatible with static-graph inference.

\section{Background on FP4 Formats}
\label{sec:background}

Four-bit floating point arithmetic is now being used to to enable highly memory- and compute-efficient inference at training at ultra-low precision. All FP4 formats discussed here follow the standard \emph{E2M1} structure: 1 sign bit, 2 exponent bits, and 1 mantissa bit. 
This FP4 format can represent the real numbers \( \mathbb{R}_{\text{E2M1}} \subset \mathbb{R} \), where
\[
\mathbb{R}_{\text{E2M1}} = \{0, \pm 0.5, \pm 1, \pm 1.5, \pm 2, \pm 3, \pm 4, \pm 6\}.
\]
\emph{Block Floating Point} formats have been widely used for quantizing DNN and Generative models as they offer an optimal tradeoff between precision, dynamic range, and hardware acceleration. The standard underlying structure for BFP formats involves a common exponent-only scale factor for a group of low-precision floating-point numbers. This common scale is responsible for projecting the unquantized numbers into the range of values representable by the low-precision format. The standard practice of this ``calibration'' computes the scale as the ratio of the maximum value present in the block and the maximum value representable by the low-precision format

The NVIDIA Blackwell architecture introduces support for two 4-bit block floating-point formats: \textbf{NVFP4} and \textbf{MXFP4}. Each of these formats represents a micro-block (i.e. a fixed-length chunk of a vector) of numbers as a vector of FP4 numbers multiplied by a single 8-bit scale factor.

\paragraph{NVFP4.} NVFP4~\cite{nvfp4} is a hardware-accelerated FP4 format that operates over \emph{micro-blocks} of 16 values, each of which shares a scale factor stored as an 8-bit E4M3 floating-point number. Let \( \mathbb{R}_{\text{UE4M3}} \subset \mathbb{R}_+ \) denote the set of positive scale values representable in that FP8 format. Then the representable blocks in NVFP4 can be denoted by the set
\[
    V_{\text{NVFP4}} = \{
        s \cdot q \mid
        q \in \mathbb{R}_{E2M1}^{16}, \, s \in \mathbb{R}_{\text{UE4M3}}
    \} \subset \mathbb{R}^{16}.
\]
A tensor \( X \in \mathbb{R}^{\cdots \times N \times D} \) (for $D$ divisible by $16$) is represented in NVFP4 by quantizing each row in non-overlapping blocks of size $16$, resulting in a pair $(Q,S)$ with
\[
Q \in \mathbb{R}_{\text{E2M1}}^{\cdots \times N \times D}, \qquad S \in \mathbb{R}_{\text{UE4M3}}^{\cdots \times N \times (D/16)}.
\]
This uses an average of $4.5$ bits per number.

\begin{figure}
\vspace{0.5em}
\begin{lstlisting}[language=Python][fontsize=\footnotesize]{cuda}
// input float SFScaleVal is global scale
// vecMax = max of abs of input vector
float SFValue = SFScaleVal * 
  (vecMax * __frcp_rz(6.0f));
__nv_fp8_e4m3 tmp = __nv_fp8_e4m3(SFValue);
float SFValue = float(tmp);
float outputScale = SFValue == 0 ? 0.0f : 
  __frcp_rz(SFValue*__frcp_rz(SFScaleVal));
// set fp2Vals = input vec * outputScale
uint32_t e2m1Vec=fp32_vec_to_e2m1(fp2Vals);
\end{lstlisting}
\vspace{-1em}
\caption{Pseudocode showing how VLLM rounds to nvfp4. This is a simplification of rounding code found in \url{vllm/csrc/quantization/fp4/nvfp4_utils.cuh}}
\label{figVLLMnvf4}
\end{figure}

To \textbf{quantize} (round) a real micro-block vector $x \in \mathbb{R}^{16}$ to NVFP4, the standard approach proceeds by first (after dividing by the global scale from row, column, and/or tensor scaling) computing a shared scale \( s \in \mathbb{R}_{\text{FP8}} \) and then the vector $q \in \mathbb{R}_{FP4}^{16}$ as
\[
    s = \operatorname{round}_{\text{UE4M3}}\left( \frac{\| x \|_{\infty}}{6.0} \right),
    \hspace{1em}
    q_i = \operatorname{round}_{\text{E2M1}}\left( \frac{x_i}{s} \right).
\]
Here, \( 6.0 \) is the largest magnitude representable in \( \mathbb{R}_{\text{E2M1}} \), $\operatorname{round}_F$ denotes ordinary nearest-neighbor rounding into the format $F$, and the infinity-norm $\| x \|_{\infty} = \max_i |x|$ is the largest absolute magnitude of an entry of $x$. 
The underlying arithmetic operations needed to do this (finding the abs-max, dividing by $6$, dividing by the scale) are usually done in FP32.
This is illustrated in the code in Figure~\ref{figVLLMnvf4}, which is thinly simplified from the VLLM codebase~\citep{kwon2023vllm}: this code is representative of the standard approach to rounding to NVFP4 in use today.

To quantize a whole tensor, the standard approach simply applies this rounding algorithm independently to each $16$-sized block of the tensor.

\textbf{Dequantization} reconstructs the full-precision tensor via
\[
\hat{x}_i = \operatorname{decode}_{\text{E2M1}}(q_i) \cdot s
\]
or, in matrix form, producing $\hat{X} \in \mathbb{R}^{N \times D}$
\[
\hat{X} = D(Q,S)=\operatorname{decode}(Q) \odot \operatorname{Broadcast}(S)
\]
where \( \operatorname{Broadcast}(S) \in \mathbb{R}^{N \times D} \) expands each \( s_g \) over its corresponding block, and \( \odot \) denotes elementwise multiplication.
But note that for typical operations, which use FP4 Tensor Cores, explicit dequantization is never performed: instead, the hardware uses the NVFP4 values direcly to compute a low-precision matrix multiplication.

\paragraph{MXFP4.} The MXFP4 format follows the same 4-bit E2M1 encoding but uses a larger group size of 32 values per block, paired with a simpler power-of-two scaling factor. Concretely, it uses a UE7M0 8-bit floating-point format instead of the UE4M3 format of NVFP4: otherwise, the quantization and dequantization algorithms are identical.
NVFP4, with its smaller block size and higher-precision FP8 scale, achieves lower quantization error and better accuracy retention in practice, but trades off for a worse dynamic range and a (very slightly) higher bits-per-number.




\section{\method}
\label{sec:scalesearch}
The standard approach of choosing the scale based only on the maximum magnitude of the input vector $x$ is obviously not optimal, in the sense that it is not guaranteed to find the vector in $V_{\text{NVFP4}}$ that is closest to $x$---but this choice of scale is generally assumed to be ``close enough'' to optimal to be a good heuristic.
Perhaps surprisingly, as we show in this section, it can be significantly suboptimal, and other scales can yield quantized vectors with much lower mean squared error, on both synthetic random vectors and real neural network data.

\begin{algorithm}[h!]
\caption{NVFP4 Scale Search}
\label{alg:scale-search}
\begin{algorithmic}
\State \textbf{Input:} Block $x \in \mathbb{R}^{16}$, search range $[f_{\min}, f_{\max}]$
\State \textbf{Output:} Best scale $s^*$, fp4 vector $q^*$, best offset $f^*$
\State $x_{\max} \gets \max_{i \in \{1,\ldots,16\}} \; | x_i |$
\State $s \gets \operatorname{round}_{\text{UE4M3}}\left( x_{\max} \cdot (1.0/6.0) \right)$ \Comment{Standard scale}
\State $s_{\text{int8}} \rightarrow \operatorname{reinterpret}(s, \text{int8})$ 
\State $\ell^* \gets +\infty$ \Comment{Initialize best loss}
\For{$f = f_{\min}$ \textbf{to} $f_{\max}$}
    \If {$1 \le s_{\text{int8}} + f \le 127$} \Comment{If scale in range}
    \State $s^{(f)} \gets \operatorname{reinterpret}(s_{\text{int8}} + f, \text{fp8}_{\text{UE4M3}})$
    \State $q_i \gets \operatorname{round}_{\text{E2M1}}\left( x_i / s^{(f)} \right), \; \forall i$ \Comment{Quantize}
    \State $\hat{x}_i \gets q_i \cdot s^{(f)}, \; \forall i$  \Comment{Dequantize}
    \State $\ell \gets \sum_{i=1}^{16} (x_i - \hat{x}_i)^2$  \Comment{Compute loss}
    \If {$\ell < \ell^*$} \Comment {Update best scale}
        \State $\ell^* \gets l$
        \State $s^* \gets s^{(f)}$
        \State $q^* \gets q$
    \EndIf
    \EndIf
\EndFor
\State \Return $s^*$, $q^*$, $f^*$
\end{algorithmic}
\end{algorithm}


We observe that the extra budget of scale mantissa bits in NVFP4 creates an opportunity for fine-grained searching of the block scale factor, rather than using the default scale based on the maximum value of the block. We implement this search mechanism by adding offsets to the default scale $s$ and selecting the scale that minimizes the quantization error of the block. We focus on the NVFP4 format to implement \method{} as it is the only format, to the best of our knowledge, that has a floating-point scale factor and is supported by modern hardware for accelerated computations. We then propose \method{}, which searches multiple FP8 scales to reduce the error of micro-block floating point quantization.

The main algorithmic idea of \method{} is to search a number of scales that are nearby the standard maximum-magnitude scale. Specifically, if $s \in \mathbb{R}_{\text{UE4M3}}$ denotes the ``standard'' scale (Section~\ref{sec:background}), we consider scales $s^{(f)}$ that are \emph{offset} from $s$ by an integer $f$, where
\[
    s^{(f)} = \operatorname{reinterpret}(\operatorname{reinterpret}(s, \text{int8}) + f, \text{fp8}_{\text{UE4M3}}).
\]
That is, $s^{(0)} = s$, $s^{(1)}$ is the smallest UE4M3 value larger than $s$, $s^{(-1)}$ is the largest UE4M3 value smaller than $s$, etc., such that
\[
    \cdots < s^{(-2)} < s^{(-1)} < s^{(0)} < s^{(1)} < s^{(2)} < \cdots.
\]
\method{} exhaustively searches all $f$ between some minimum $f_{\min}$ and maximum $f_{\max}$ for the search that minimizes the (mean squared) quantization error.
The full algorithm for NVFP4 is presented in Algorithm~\ref{alg:scale-search}; this algorithm can easily be modified to handle other micro-block floating point formats.

Obviously, if we let $f_{\min} = -127$ and $f_{\max} = +127$, searching exhaustively over all scales, then Algorithm~\ref{alg:scale-search} would be guaranteed to find the closest representable vector to the input $x$. But this would be computationally costly. In the next subsection, we will show via experiments on synthetic and real data that searching only a small subset of scale offsets suffices to get a substantial reduction in error and come very close to the optimal MSE.


\begin{figure}[!t]
\centering
\includegraphics[width=0.9\linewidth]{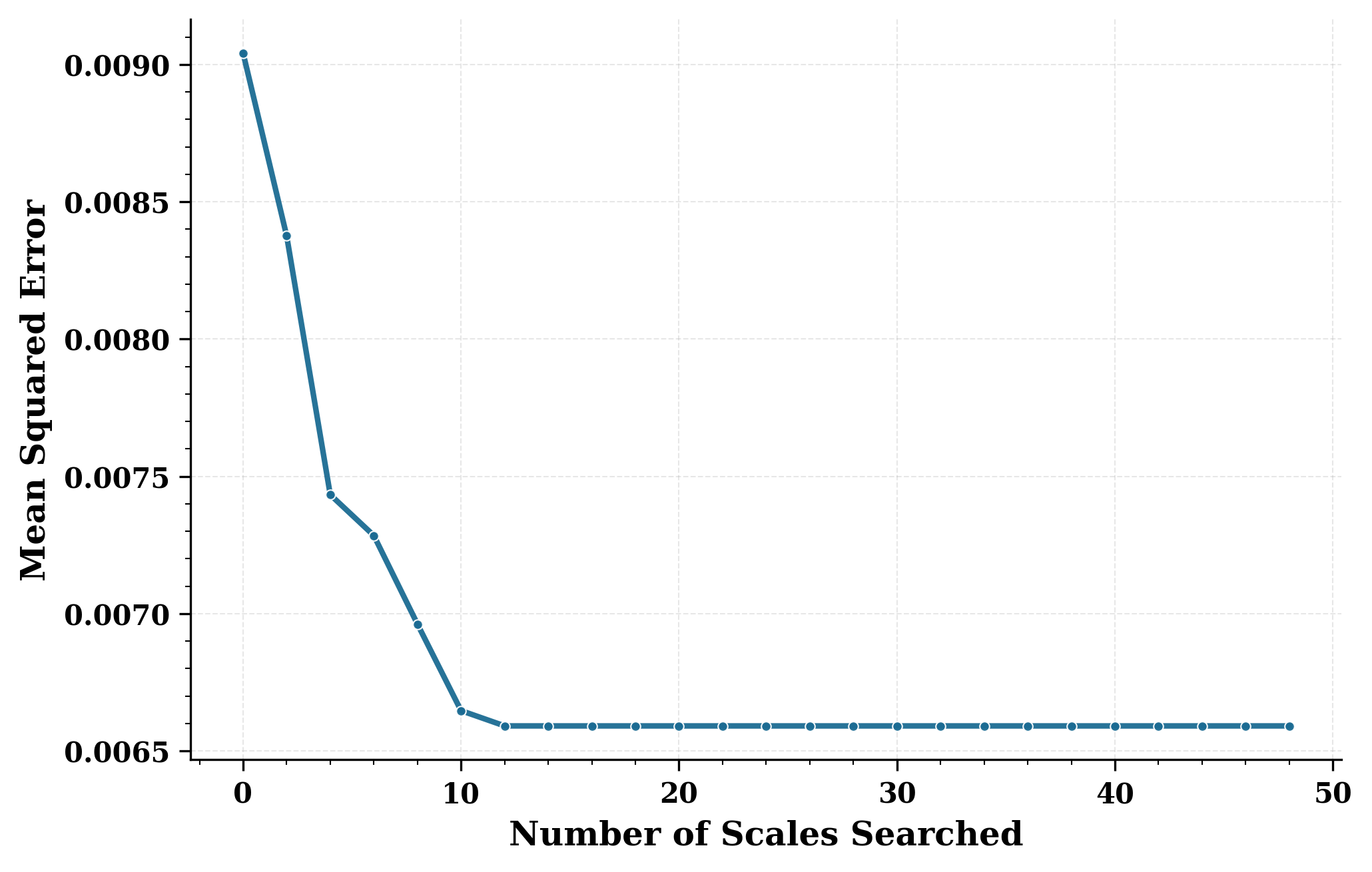} 
\caption{Quantization MSE for unit Gaussian tensor}
\label{fig:mse}
\end{figure}

\begin{figure}[!t]
\centering
\includegraphics[width=0.9\linewidth]{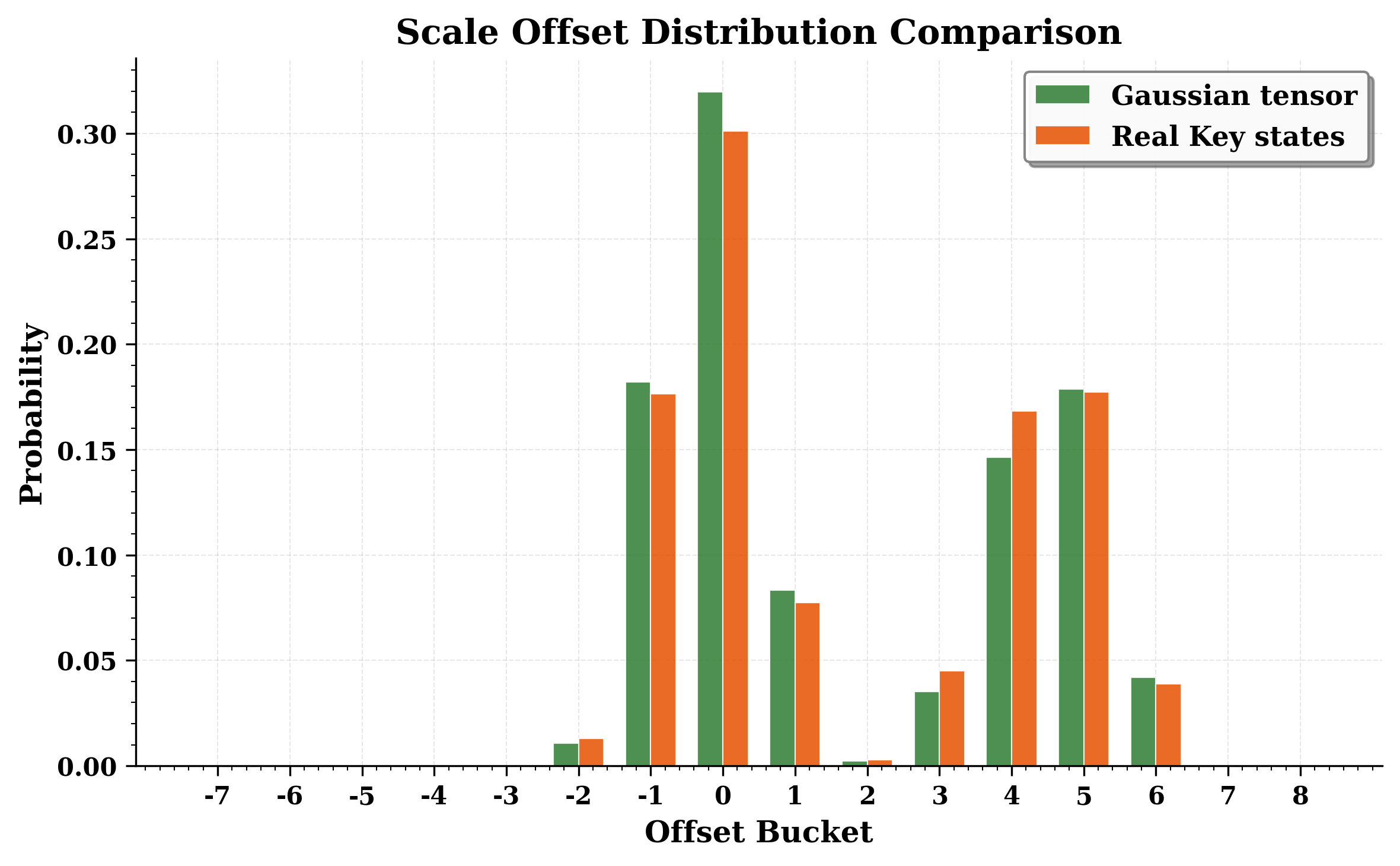}
\caption{Offset distribution for Gaussian and real key states tensor for nvfp4 format.}
\label{fig:histogram}
\end{figure}

\begin{figure*}[!t]
\centering
\begin{subfigure}{0.32\textwidth}
    \centering
    \includegraphics[width=\linewidth]{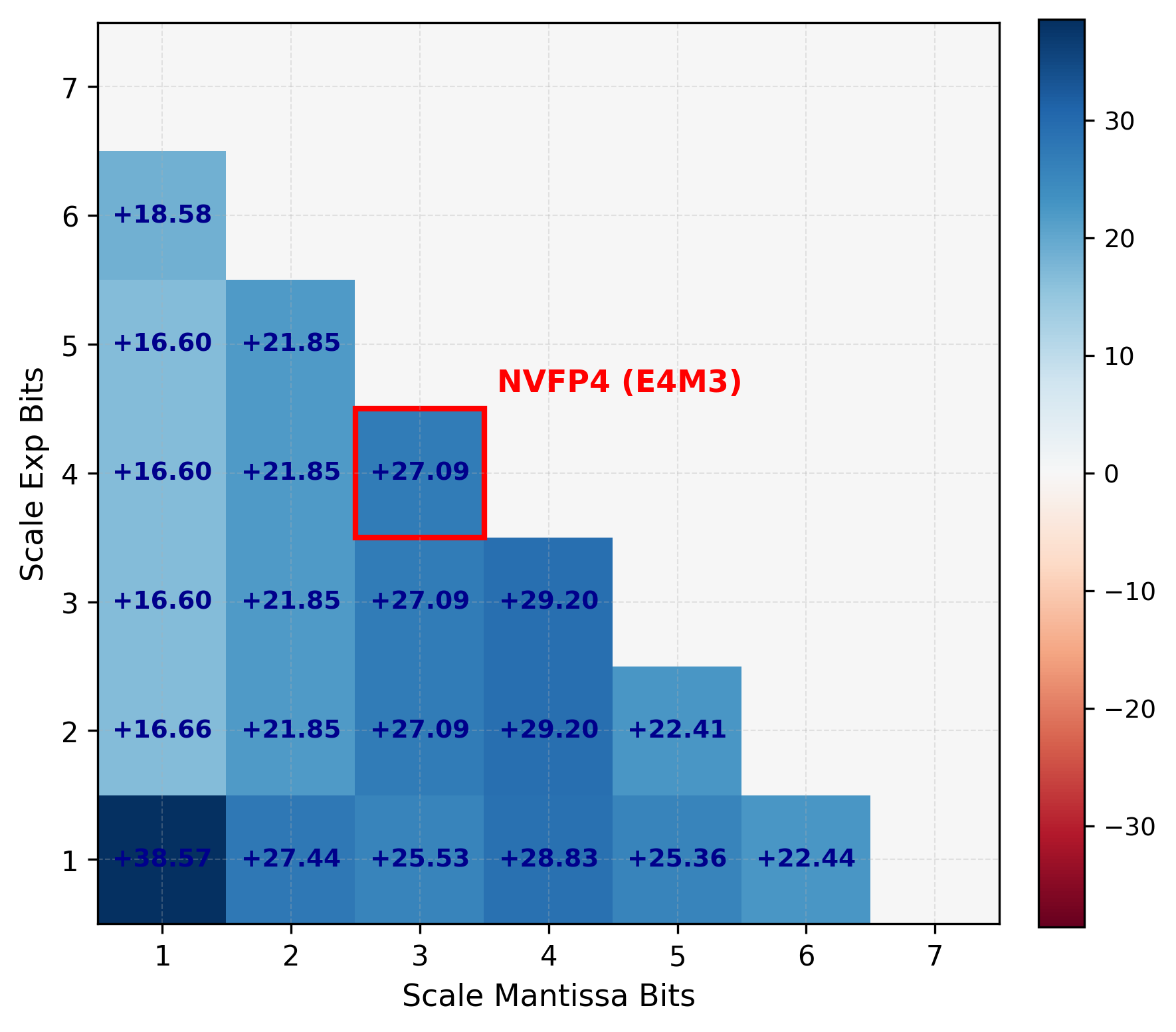}
    \caption{}
    \label{fig:nvfp-scale}
\end{subfigure}
\hfill
\begin{subfigure}{0.32\textwidth}
    \centering
    \includegraphics[width=\linewidth]{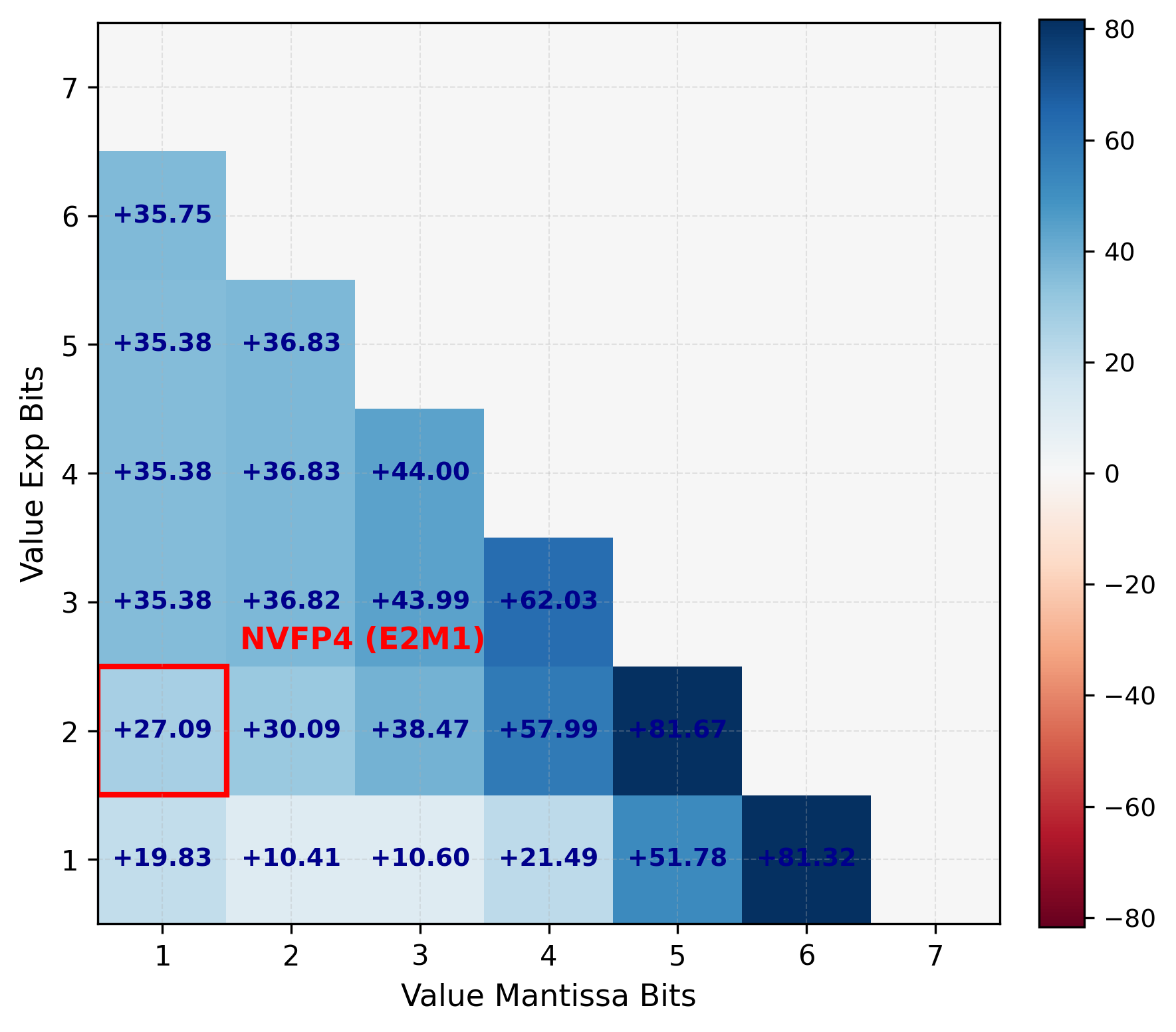}
    \caption{}
    \label{fig:nvfp-val}
\end{subfigure}
\hfill
\begin{subfigure}{0.32\textwidth}
    \centering
    \includegraphics[width=\linewidth]{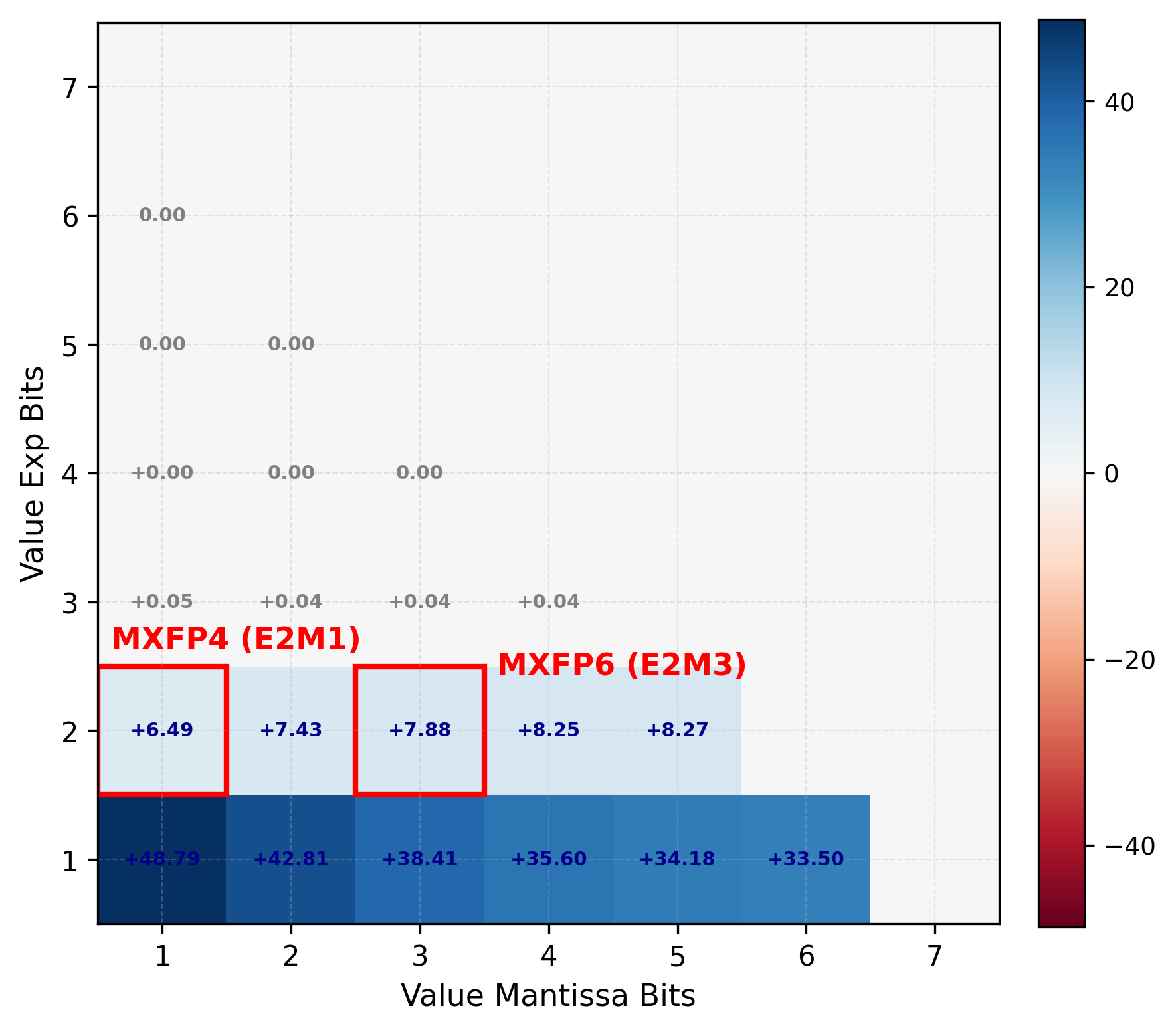}
    \caption{}
    \label{fig:mxfp}
\end{subfigure}
\caption{Simulated percentage improvement by \method for different scale and value configurations. (a) Scale representation sweep with value format fixed at E2M1. (b) Value representation sweep with scale format fixed at E4M3. (c) MXFP value representation sweep with scale format fixed at E8M0. Standard formats are marked in red.}
\end{figure*}

\begin{figure}[!h]
\centering
\includegraphics[width=0.9\linewidth]{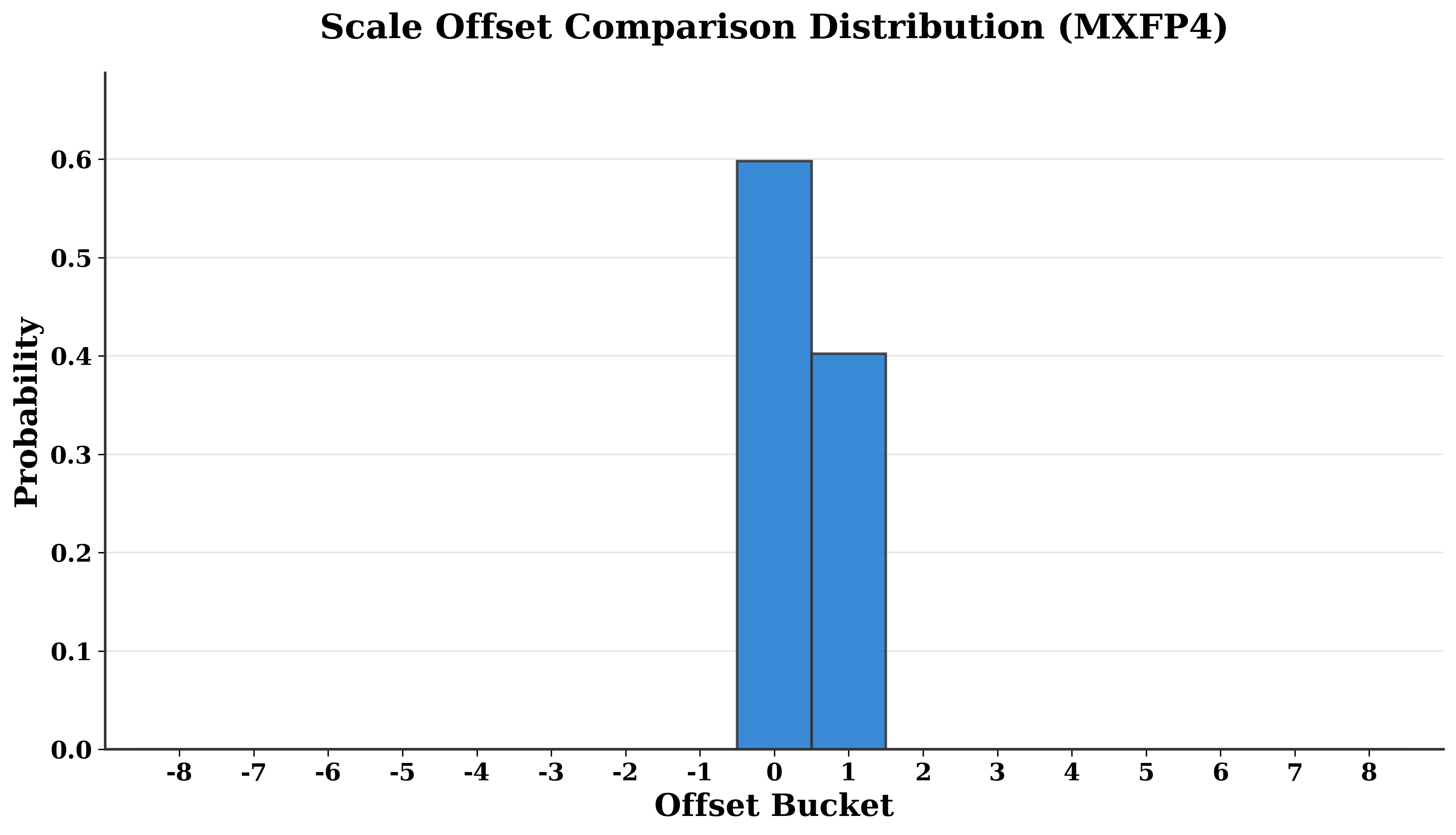}
\caption{Offset distribution for Gaussian data for mxfp4 format.}
\label{fig:histogrammxf4}
\end{figure}


\begin{figure}[!t]
\centering
\includegraphics[width=0.5\textwidth]{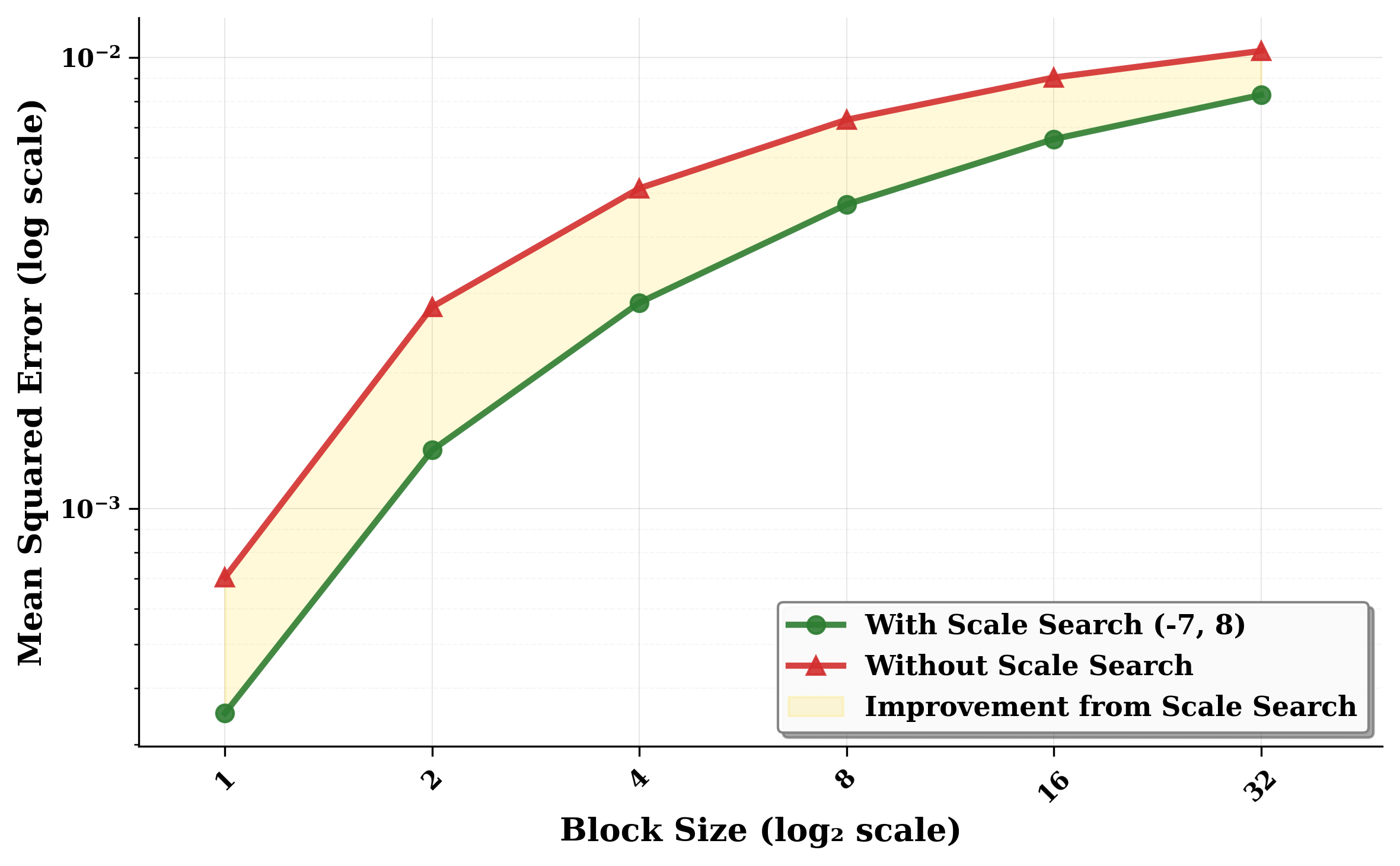}
\caption{\method advantage reduces as the block size increases.}
\label{fig:block_size}
\end{figure}


\subsection{Synthetic Validation}
We first test our approach on synthetic data by generating a large FP32 tensor with values sampled from a standard Gaussian distribution and quantizing it to NVFP4 with Algorithm~\ref{alg:scale-search} across a range of different numbers of scales searches (the number of scales searched is $f_{\max} - f_{\min} + 1$, and we chose ranges where $f_{\min} = 1-f_{\max}$). \cref{fig:mse} shows the quantization mean squared error (MSE) for this tensor: as the number of scales searched is increased, the quantization error reduces, until it saturates without any further reduction with an increase in search scope. On these synthetic data, the MSE went down from $0.0990$ to $0.0066$, an reduction by about $25\%$.

\paragraph{Offset distribution}

Motivated by a need to reduce the range of the search, we look at the distribution of offsets chosen for different blocks. Figure \ref{fig:histogram} shows the empirical distribution of offsets resulting from a exhausive search. The distribution has a bimodal structure with two curves centered around offsets $0$ and $4$ for a generated Gaussian-sampled tensor. The figure also plots the same distribution for a real Key state tensor taken from Llama 3.1 8B model, showing that this behavior persists even when quantizing real data. This suggests that the scale search analysis done on a Gaussian-sampled tensor can be safely applied for the quantization of real model tensors, owing to the similar offset distribution as seen in \cref{fig:histogram}. Based on this empirical analysis, we choose an offset set range of $f_{\min} = -2$ to $f_{\max} = +6$ while searching for the optimal scale in our real model quantization pipeline: we use this search range for the remainder of the NVFP4 experiments in this paper.

The intuition for the observed distribution is that the “default” block scale (offset 0) is chosen such that the largest-magnitude entry will have low error when represented with the largest-magnitude fp4 value, 6 (or -6). The popularity of offsets near 0 corresponds to quantizations where the largest-magnitude entry will be represented by 6 (or -6). But another way the largest-magnitude entry can be stored in fp4 is as 4 (or -4), the second-largest-magnitude fp4 value. 

Here, we want the scale to be about $1.5x = 6/4$ larger than the “default” block scale, to account for the fact that the largest-magnitude entry is stored as 4 rather than 6. An offset of around 4 or 5 typically corresponds to about a $1.5x$ factor: to illustrate, the e4m3 number 1.0 corresponds to the bit pattern 0 1000 000 = 64, and adding an offset of 4 to this yields 0 1000 100 = 68, which corresponds to the number 1.5. The mode of the distribution around 4\&5 in Figure \ref{fig:histogram}. corresponds to quantizations where the largest-magnitude entry will be represented by 4 (or -4). There is no “third mode” for representing the largest-magnitude entry with 3 (or -3) because this cannot do better than representing it with 6 (or -6) and halving the scale.

The above considerations are generic, rather than depending on any particular data distribution. And this is validated by our empirical observation that the offset distribution has a mode presented in Figure \ref{fig:histogram}, which emerges for both synthetic Gaussian tensors and real activation tensors.

\paragraph{Other formats}

ScaleSearch can be used with any block quantization format, with varying benefits depending on the scale and value format used. We simulated ScaleSearch improvements for a range of configurations with varying number of exponent and mantissa bits for scale and value formats. Figure \ref{fig:nvfp-scale} shows the percentage improvement realised by \method for hypothetical block quantization formats having different scale representation bits with value format fixed at E2M1 (same as in NVFP4). Figure \ref{fig:nvfp-val} shows the same analysis for different value representation bits with scale format fixed as E4M3 (same as in NVFP4). \method reduces the quantization error upto about 80\%, with a reduction of 27\% in particular for NVFP4 format. We also study the effect of \method for the standardised MXFP format (Figure \ref{fig:mxfp}), and we observe a reduction of quantization MSE of 11\% for MXFP6 with E2M3 values and 8\% for MXFP4.

We further note that \method has better performance for recent low-precision formats such as NVFP4 for two reasons.
First, we observe that the benefit of \method is more pronounced with smaller block sizes.
Figure \ref{fig:block_size} shows that as the block size increases, the MSE gap between search and no-search cases reduces. Prior quantization methods had larger block sizes ranging from per column/row to per-tensor scaling, wherein the benefit of \method would be diminished. Second, as Figure~\ref{fig:histogram} shows, it is most profitable to search scales near the max-abs-scale $s$, and the representable scale values in MXFP4 are simply spread out more over a larger range, placing fewer of them ``close'' to $s$. 
This behavior can be seen explicitly in Figure~\ref{fig:histogrammxf4}, which plots the same distribution for MXFP4 scaling: observe that only two offsets are ever used and the most common choice is no offset.


\subsection{Application to language modelling}

We extend \method to develop \methodattention, an end-to-end attention method which optimizes attention computation for inference by formulating the problem into native NVFP4 format with a hardware-aware pipeline void of any de-quantization overheads. All tensors involved in the attention layer ($\mathbf{Q,K,P,V}$) are quantized to the NVFP4 format and directly multiplied using the NVFP4 Tensor Cores with a FP32 accumulator, enabling higher throughput. Observe that, in addition to any compute benefits, this reduces the memory footprint of the KV cache by storing it in NVFP4 format (4.5 bits). Block scales for ($\mathbf{Q,K,P,V}$) are computed using \method after per-row scaling, and quantization is done along the reduction dimension of a matrix multiply, which is constrained by the requirements of the NVFP4 MMA instruction \cite{scale_dims}. To close the accuracy gap with respect to unquantized models, \methodattention integrates two additional techniques on top of \method. 


 

    

\begin{figure*}[!t]
\centering
\includegraphics[width=0.85\textwidth]{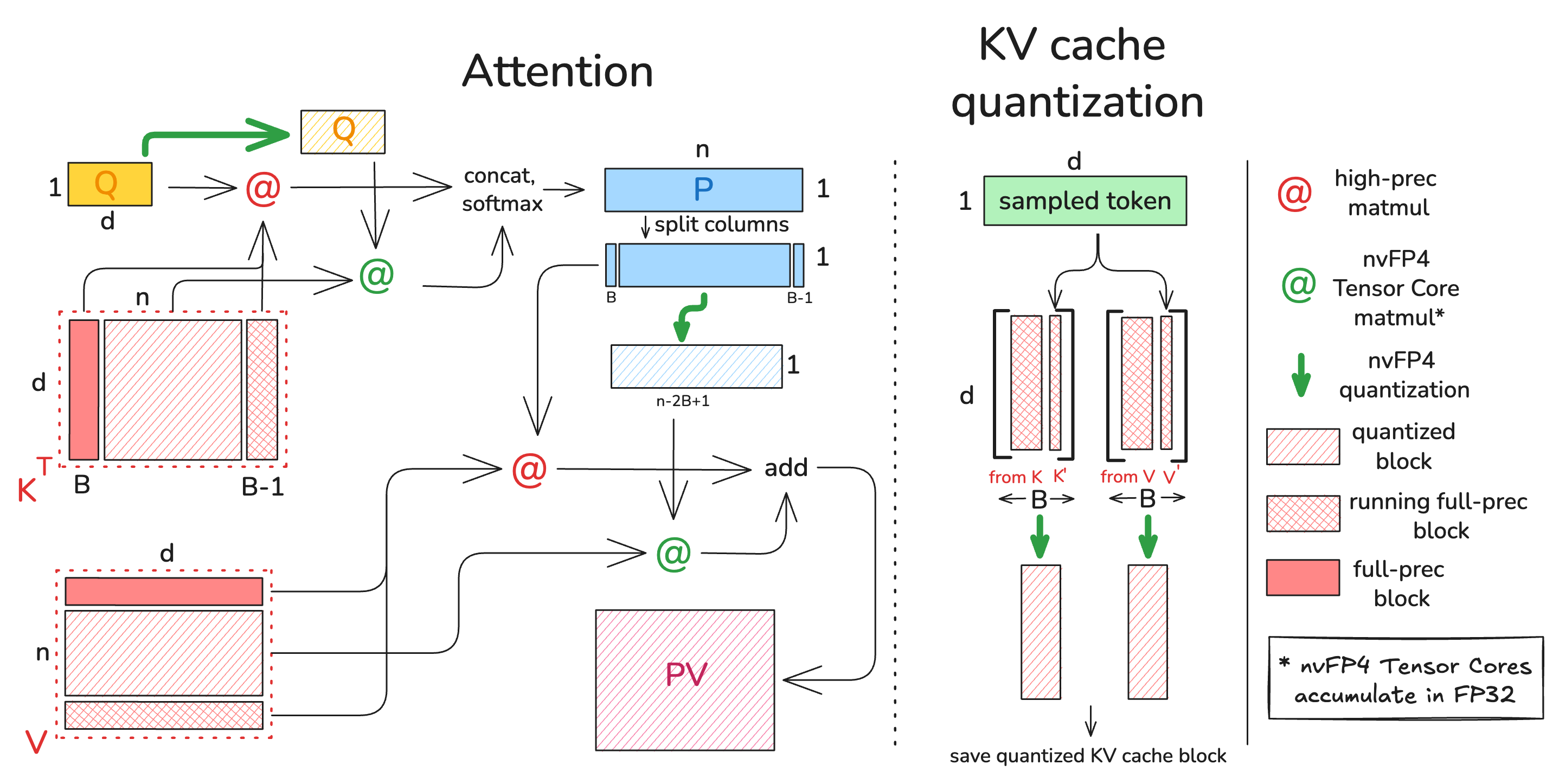}
\caption{\methodattention workflow example for inference, where $n$ tokens (such that $n \mod B=B-1$) have been processed. Mixed precision K is multiplied with Q using a majority of nvFP4 Tensor Core instructions, which accumulates P in FP32. P is further quantized and undergoes a mixed-precision multiply with V. The Key, Value states corresponding to the new sampled token completes the block of size B, which is then sent for quantization and stored in the compressed KV cache.}
\label{fig:ssa}
\end{figure*}

\paragraph{Incoherence Processing with magnitude reduction}
Multiple prior works like QuIP \cite{quip}, QuIP\# \cite{quipsharp}, QuaRot \cite{quarot} have proposed using Incoherence Processing (IP) as a principled way of reducing outliers in the tensor to be quantized. Following QuIP\# \cite{quipsharp}, we use a Hadamard matrix $H$ to transform $Q$ and $K$ matrices, in a way that preserves the attention scores, while reducing quantization error. On top of traditional IP, we implement an additional transformation that reduces the average squared magnitude of the projected query ($Q$) and key ($K$) representations, as this directly reduces the quantization error. Specifically, we introduce a pair of linear transformations:
\[
Q' = Q R^{-T}, \qquad K' = K R,
\]
where $R \in \mathbb{R}^{d \times d}$ is an invertible transformation. 
This transformation preserves the attention scores, while reducing the quantization error for query ($Q$) and key ($K$):
\[
Q' {K'}^\top = (Q R^{-T})(K R)^\top = Q R^{-T} R^\top K^\top = Q K^\top.
\]
A derivation of these formulas, along with a proof that they in some sense minimize the averaged squared magnitude of the projected $Q$ and $K$ matrices, appears in the Appendix - we omit it here both for brevity and as it is in some sense orthogonal to our main contribution of \method.


\paragraph{Attention-sink-aware mixed-precision cache}
Attention sink analysis \cite{spa5} presented that attention scores are concentrated towards the most recent ``local'' tokens and initial tokens of the context. Following this observation, some previous KV cache compression methods like KVQuant \cite{kvquant} and IntactKV \cite{intactkv} preserve the KV cache corresponding to initial and ``pivot'' tokens in high precision. Expanding on this idea, \methodattention uses mixed-precision attention computation, wherein the entire attention matrix ($QK^T$) is divided into blocks of size $B$ and the first and last (most recent) blocks are computed in high precision, while all other blocks are computed in lower precision. This is implemented by storing a constant $O(B)$ sized KV cache in unquantized form. This does not scale with context or generation length, and hence does not make inference more memory-bound. Block size $B$ can modified with respect to performance-optimality for given memory and compute resources. The only requirement is $m\geq16$ as $V$ is quantized along the token dimension and quantization is performed in groups of 16 when using the NVFP4 format. 

\paragraph{\methodattention{} workflow}
Figure \ref{fig:ssa} presents the implementation and workflow for \methodattention, with an inference example in the prefill stage where $n$ tokens (such that $n \% B=B-1$) have been processed. Mixed-precision KV cache corresponding to these $n$ tokens contains the first block and the incomplete last block (containing $B-1$ tokens)in full-precision, while rest of the KV cache is in the NVFP4 format. Quantized K is multiplied with quantized Q using a nvFP4 Tensor Core instructions, while the full precision matrix multiplications are done with high-precision matmul instructions. Since unquantized values do not scale with context length, majority of the computations are performed using the fast nvFP4 Tensor Cores, which accumulates the results in FP32. Results from the two matrix multiply operations are concatenated to compose $QK^{T}$, which is used to calculate probabilities using the softmax operation. P matrix is further quantized and undergoes a mixed-precision multiply with V (similar to $QK^{T}$) - the only difference being that V matrix is partitioned along the rows, and hence P matrix needs to be split along columns to perform the partial high and low precision matrix multiply, which is then added together to get the $PV$ matrix. The Key, Value states corresponding to the new sampled token are concatenated at the end of the incomplete unquantized K and V blocks respectively, and the complete blocks are then quantized into nvFP4 and stored as compressed KV cache. Since all matrices are quantized and nvFP4 Tensor Cores accumulate in FP32, there is no dequantization step involved.

\section{Experiments}
\label{sec:scale_search_integration}

We evaluate the impact of using \method by validating the following claims about \method 's improved numerical stability and minimal overhead. First, we evaluate the improved quality of \method in the context of PTQ of language models and ultra-low-precision attention for diffusion models inference. Furthermore, we benchmark the throughput of \method in comparison to other low-precision attention alternatives to illustrate the minimal overhead of \method. Finally, we evaluate \methodattention for language models.

\subsection{\method}
\label{sec:attention}

\begin{table*}[!t]
\centering
\caption{\method as a PTQ technique evaluated across several general capability benchmarks averaged across 5 runs (MMLU is run once). Higher is better, and \textbf{bold} is best. }
\label{tab:results}
\begin{small}
\begin{tabular}{clcccc}
\toprule
\textbf{Model} & \textbf{Method} & \textbf{GPQA} & \textbf{MATH-500} & \textbf{AIME-120} & \textbf{MMLU}  \\
\midrule
\multirow{3}{*}{%
  \rotatebox[origin=c]{0}{\shortstack{\bfseries DeepSeek-R1-\\ \bfseries Distill-Qwen-1.5B}}
}
 & Baseline & 32.6 {\scriptsize (3.2)} & 64.6 {\scriptsize (1.4)} & 24.5 {\scriptsize (2.3)} & 48.0  \\
\cmidrule(lr){2-6}
 & NVFP4    & 30.4 {\scriptsize (4.2)} & 51.6 {\scriptsize (2.9)} & 18.7 {\scriptsize (1.4)} & 45.2 \\
 & \methodbold  & \textbf{31.3} {\scriptsize (1.8)} & \textbf{62.1} {\scriptsize (1.9)} & \textbf{19.8} {\scriptsize (1.9)} & \textbf{45.4}  \\
\midrule
\multirow{3}{*}{%
  \rotatebox[origin=c]{0}{\shortstack{\bfseries Qwen3-8B}}
}
 & Baseline & 51.3 {\scriptsize (3.8)} & 72.8 {\scriptsize (3.4)} & 71.0 {\scriptsize (3.0)} & 79.7  \\
\cmidrule(lr){2-6}
 & NVFP4    & 42.4 {\scriptsize (5.5)} & 73.1 {\scriptsize (4.3)} & \textbf{63.7} {\scriptsize (1.3)} & 77.7  \\
 & \methodbold  & \textbf{49.9} {\scriptsize (1.6)} & \textbf{88.1} {\scriptsize (0.6)} & 63.0 {\scriptsize (2.2)} & \textbf{79.4}  \\
\bottomrule
\end{tabular}
\end{small}
\end{table*}

\paragraph{Post-Training Quantization}

We evaluate the use of \method as a technique for offline Post-Training Quantization (PTQ) of DeepSeek-R1-Distill-Qwen-1.5B~\cite{guo2025deepseek} and Qwen3-8B~\cite{qwen3}. We compare against the base model and an NVFP4 model quantized using TensorRT-Model-Optimizer~\cite{trt_opt} (ModelOpt). For \method we modify the ModelOpt NVFP4 quantization path to conduct \method. We utilize five general capability benchmarks (GPQA~\cite{gpqa}, MATH-500~\cite{math500}, AIME-120~\cite{aime_2024}, MMLU~\cite{mmlu}). 
GPQA tests graduate-level scientific reasoning and multi-hop inference, highlighting quantization’s impact on domain knowledge. AIME and MATH-500 focus on high-school math, stressing symbolic reasoning and precision. MMLU spans diverse subjects like history and law, evaluating world knowledge and problem-solving.  
The results are presented in Table \ref{tab:results}.

We observe \method outperforms NVFP4 across all benchmarks by up to 15 percentage points. In the only instance where NVFP4 is better (AIME-120 on Qwen3-8B), \method is still within one standard deviation (which we believe is due to the randomness inherent within these benchmarks). Notably, we see \method close the gap on benchmarks where there is a significant difference between the non-quantized baseline and NVFP4 (10.5 on MATH-500, 7.5 on GPQA, and 1.7 on MMLU). These trends are consistent across model sizes. 

\begin{table*}[!t]
\centering
\caption{Mochi and CogVideoX quality (VQA-a, VQA-t, FScore) and video-caption alignment (CLIPSIM, CLIP-t) using SageAttention3 and SageAttention3 + \method on the SageAttention3 evaluation dataset~\cite{sageattention3}. Higher is better, and \textbf{bold} is best.}
\label{tab:text2vido_quality_performance}
\begin{small}
\begin{tabular}{ccccccc}
\toprule
\textbf{Model} & \textbf{Method} & \textbf{VQA-a} $\uparrow$ & \textbf{VQA-t} $\uparrow$ & \textbf{FScore} $\uparrow$ & \textbf{CLIPSIM} $\uparrow$ & \textbf{CLIP-T} $\uparrow$ \\
\midrule
\multirow{3}{*}{Mochi}
 & Full-Precision & 62.919 & 70.216 & 2.466 & 0.1834 & 0.9991 \\
\cmidrule(lr){2-7}
 & SageAttention3 & 49.635 & 55.738 & 1.874 & \textbf{0.1829} & 0.9988 \\
 & SageAttention3 + \methodbold & \textbf{57.424} & \textbf{70.362} & \textbf{2.143} & 0.1826 & \textbf{0.9989} \\
\midrule
\multirow{3}{*}{CogvideoX}
 & Full-Precision & 74.969 & 75.345 & 5.969 & 0.1850 & 0.9976 \\
\cmidrule(lr){2-7}
 & SageAttention3 & 72.915 & 74.912 & 5.118 & \textbf{0.1845} & 0.9973 \\
 & SageAttention3 + \methodbold & \textbf{75.753} & \textbf{75.873} & \textbf{5.402} & 0.1833 & \textbf{0.9974} \\
\bottomrule
\end{tabular}
\end{small}
\end{table*}

\paragraph{Diffusion Inference attention}
\label{sec:experiments-quality}
We additionally evaluate the quality of \method in combination with SageAttention3 \cite{sageattention3} for end-to-end video diffusion model inference. We compare SageAttention3 \cite{sageattention3} with and without \method in Mochi \cite{genmo2024mochi} and CogvideoX-2B \cite{yang2024cogvideox}, using CLIPSIM and CLIP-Temp (CLIP-T) to measure text-to-video alignment and VQA-a, VQA-t, and FScore to measure quality and consistency. We evaluate these models on the SageAttention3 evaluation dataset~\cite{sageattention3}.

As seen in \cref{tab:text2vido_quality_performance}, \method improves quality for Mochi and CogVideoX over SageAttention3 in VQA-a, VQA-t, and FScore, while matching in CLIPSIM and CLIP-T. Similar to our PTQ results, we observe that \method matches naive SageAttention3 in metrics where SageAttention3 is close to full-precision attention (e.g., CLIPSIM and CLIP-T). However, we find that in metrics where SageAttention3 struggles in comparison to full-precision attention, incorporating \method significantly improves quality  (e.g., VQA-a, VQA-t, FScore).

\begin{table*}[!t]
\caption{Perplexity evaluation on Wikitext-2 test set. Lower is better, and \textbf{bold} is best.}
\label{tab:ppl}
\begin{center}
\begin{small}
\begin{sc}
\begin{tabular}{lrrrr}
\toprule
\textbf{Method} & \textbf{Llama 3.1 8B} & \textbf{Llama 3.1 70B} & \textbf{Qwen3 4B} & \textbf{Qwen3 8B} \\
\midrule
FullPrec & 5.4837 & 2.5554 & 11.1327 & 8.3013 \\
\midrule
naive-FP4 & 5.9988 & 3.4000 & 11.5258 & 8.4429 \\
naive-FP4 + \method & 5.8330 & 3.3441 & 11.3618 & 8.4378 \\
SA3 & 5.9542 & 3.3899 & 11.3672 & 8.4575 \\
SA3 + \method & 5.8060 & 3.2972 & 11.2441 & 8.4181 \\
\methodattention & \textbf{5.4977} & \textbf{2.6348} & \textbf{11.2088} & \textbf{8.3018} \\
\bottomrule
\end{tabular}
\end{sc}
\end{small}
\end{center}
\vskip -0.1in
\end{table*}

\begin{table}[!t]
\caption{Language benchmark evaluation on GPQA:diamond for Llama 3.1 8B Instruct model. \textbf{Bold} is best.}
\label{tab:gpqa}
\begin{center}
\begin{small}
\begin{sc}
\begin{tabular}{lr}
\toprule
\textbf{Method} & \textbf{Accuracy} \\
\midrule
FullPrec & 31.81 \\
\midrule
SA3 & 26.26 \\
\methodattention & \textbf{32.32} \\
\bottomrule
\end{tabular}
\end{sc}
\end{small}
\end{center}
\vskip -0.1in
\end{table}

\begin{table}[!t]
\caption{Ablation analysis of \methodattention (SSA) variants. Note : w/o refer to without.}
\centering

\begin{tabular}{l c}
\toprule
\textbf{Method} & \textbf{PPL} \\
\midrule
SSA & 5.4977 \\
SSA w/o ScaleSearch & 5.5024 \\
SSA w/o IP and Magnitude Reduction & 5.5283 \\
SSA w/o Mixed-precision KV cache & 5.5768 \\
\bottomrule
\end{tabular}
\label{tab:ablation}
\end{table}

\subsection{\methodattention}
\label{sec:swiftattention-eval}
In this section, we evaluate \methodattention using simulated quantization framework developed on top of Pytorch \cite{PyTorch}. We test task-independent PPL of quantized and unquantized models.

\paragraph{Perplexity}
We use  Llama 3.1 8B, Llama 3.1 70B \cite{llama3}, Qwen3 4B, and Qwen3 8B \cite{qwen3} for measuring token perplexity on the test set of Wikitext-2 dataset \cite{Wikitext-2}. For PPL evaluation, we compare against the full native precision of the respective model (FullPrec), the vanilla FP4 quantization (Naive-FP4) \cite{nvfp4}, and SageAttention3 (SA3) \cite{sageattention3}. For SageAttention3, we simulate the given algorithm (Algorithm 1) using our FP4 simulator as the provided code is numerically unstable when used in a causal setting. Table \ref{tab:ppl} presents these results. \methodattention outperforms SageAttention3 and Naive-FP4 across all models, and reduces PPL by upto 22\% for Llama 3.1 70B model. Note that \methodattention provides significant improvement for models of larger sizes as well, for which quantization generally produces less effect \cite{reasoninghurtsquant}.  
We use perplexity evaluation to also study the effect of \method by adding it on top of naive NVFP4 (naive-FP4) quantization and SageAttention3(SA3). Naive-FP4 involves simulated quantization of $Q,K,V,P$ using the algorithm provided by NVIDIA \cite{nvfp4}. Adding \method improves the PPL for both SageAttention3 and Naive-FP4, validating its effectiveness and applicability across a wide set of quantization algorithms. 

\paragraph{Language benchmark}  We further evaluate our method on the GPQA Diamond \cite{gpqa} benchmark using the Llama 3.1 8B Instruct model. Consistent with the perplexity results, \methodattention achieves the best performance, attaining an accuracy of 32.32, outperforming SA3 (26.26), and matching the full precision evaluation.

\paragraph{\methodattention ablation} We also study the importance of each component of \methodattention in the ablation analysis presented in \ref{tab:ablation}. The full \methodattention configuration achieves the lowest perplexity (5.4977), indicating the effectiveness of jointly combining all proposed components. Removing ScaleSearch leads to a noticeable degradation (5.5024), highlighting the importance of optimal microscaling. Similarly, disabling importance preservation (IP) and magnitude reduction increases perplexity further (5.5283), while the largest drop in performance is observed when mixed-precision KV cache is removed (5.5768), underscoring its critical contribution to preserving attention quality.

\subsection{Overhead and Efficiency}

In this section, we analyze the performance of \method by evaluating (i) quantization overhead under different search ranges, (ii) attention throughput, and (iii) end-to-end generation latency.

\paragraph{Quantization overhead.}
We first measure the overhead introduced by \method during FP32 to NVFP4 quantization. The experiment is conducted on a random Gaussian matrix of size $2048 \times 2048$, using vLLM's quantization implementation as the baseline, upon which \method is integrated. As shown in \ref{tab:quant_overhead}, the baseline quantization takes 0.0258 ms, while incorporating \method with a restricted search range $[-1,1]$ increases the time to 0.0328 ms (1.27$\times$ overhead). Expanding the search to the full range $[-2,6]$ results in 0.0449 ms (1.74$\times$ overhead). This indicates that \method introduces minimal practical overhead during quantization while providing consistent improvements in quantization MSE.

\begin{table}[t]
\centering
\caption{Quantization overhead of \method under different search ranges.}
\begin{tabular}{l c c}
\toprule
\textbf{Method} & \textbf{Time (ms)} & \textbf{Overhead} \\
\midrule
FP32 $\rightarrow$ NVFP4 (baseline) & 0.0258 & 1.00$\times$ \\
+ ScaleSearch ($[-1,1]$) & 0.0328 & 1.27$\times$ \\
+ ScaleSearch ($[-2,6]$) & 0.0449 & 1.74$\times$ \\
\bottomrule
\end{tabular}
\label{tab:quant_overhead}
\end{table}

\begin{figure*}[!t]
    \centering
    \includegraphics[width=\textwidth]{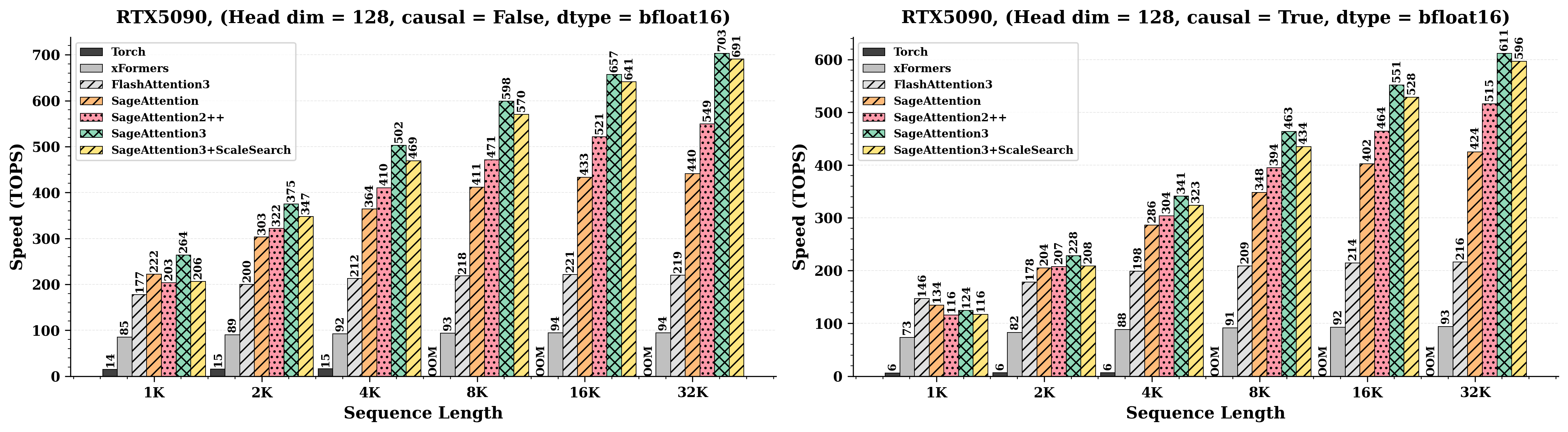}
    \caption{We benchmark the combination of SageAttention3 \cite{sageattention3} and \method against other ultra-low precision techniques \cite{sageattention1,sageattention2}, flash-attention \cite{fa3}, xformers \cite{xFormers2022}, and eager torch given an input of data type bfloat16. We find the \method doesn't significantly impact the throughput of both causal and non-causal attention, especially at larger sequence lengths.}
    \label{fig:sageattention_perf_5090_hd_128_dtype_bf16}
\end{figure*}

\paragraph{Attention throughput.}
Next, we evaluate the impact of \method on attention throughput in an experimental setup inspired by SageAttention3 \cite{sageattention3}. As shown in \cref{fig:sageattention_perf_5090_hd_128_dtype_bf16}, combining \method with SageAttention3 achieves throughput nearly identical to the baseline across both causal and non-causal settings. In particular, at long sequence lengths (32K), \method attains up to 98.3\% of baseline TOPs in non-causal attention and 97.5\% in causal attention. This demonstrates that the additional scaling search does not introduce meaningful runtime overhead in the critical attention kernel, especially in the high-throughput regime.

\paragraph{End-to-end latency.}
Finally, we measure end-to-end latency on text-to-video generation models. As shown in \cref{tab:text2video_e2e_latency}, both SageAttention3 and SageAttention3 + \method significantly reduce latency compared to full-precision attention. Importantly, \method introduces only a marginal increase over SageAttention3 (e.g., 353.40 s vs.\ 364.68 s for Mochi, and 61.72 s vs.\ 63.09 s for CogvideoX), confirming that its overhead is minimal in realistic generation workloads. Overall, these results demonstrate that \method achieves improved numerical performance with negligible impact on system efficiency.

\begin{table}[!t]
    \centering
    \caption{We report the average end-to-end latency in seconds of text-to-video generation models Mochi \cite{genmo2024mochi} and CogvideoX \cite{yang2024cogvideox} on tweleve different prompts, using full-precision (blfoat16) attention, SageAttention3 \cite{sageattention3}, and SageAttention3 with \method.
    Lower is better. \textbf{Bold} is best, and \textunderscore{underline} is second best.}
    \begin{tabular}{ccc}
    \toprule
         \textbf{Model} & \textbf{Method} & \textbf{E2E Latency} \\
         \midrule
         \multirow{3}{*}{Mochi} & Full-Precision & 503.38 \\
                                   & SageAttention3 &  \textbf{353.40} \\
                                   & SA3 + \methodbold &  \underline{364.68} \\
                                   \midrule
         \multirow{3}{*}{CogvideoX} & Full-Precision &  91.67 \\
                                   & SageAttention3 &  \textbf{61.72} \\
                                   & SA3 + \methodbold &  \underline{63.09} \\
                                   \bottomrule
    \end{tabular}
    \label{tab:text2video_e2e_latency}
\end{table}

\section{Conclusion}
In this paper, we introduce \method, a technique for selecting micro-block scaling values for nvFP4 quantization.
We find that \method reduces quantization error, leading to improvements in end-to-end quality for language modeling and video diffusion.
As an extension, we propose \methodattention, a first-of-its-kind attention algorithm for nvFP4 quantization of the KV cache using \method.
We hope these methods can help drive further development and innovation for quantization.

\bibliography{example_paper}

\begin{thebibliography}{72}
\providecommand{\natexlab}[1]{#1}
\providecommand{\url}[1]{\texttt{#1}}
\expandafter\ifx\csname urlstyle\endcsname\relax
  \providecommand{\doi}[1]{doi: #1}\else
  \providecommand{\doi}{doi: \begingroup \urlstyle{rm}\Url}\fi

\bibitem[AI \& vLLM Project(2024)AI and vLLM Project]{llmcompressor}
AI, R.~H. and vLLM Project.
\newblock {LLM Compressor}, 8 2024.
\newblock URL \url{https://github.com/vllm-project/llm-compressor}.

\bibitem[Ainslie et~al.(2023)Ainslie, Lee-Thorp, de~Jong, Zemlyanskiy, Lebrón, and Sanghai]{gqa}
Ainslie, J., Lee-Thorp, J., de~Jong, M., Zemlyanskiy, Y., Lebrón, F., and Sanghai, S.
\newblock Gqa: Training generalized multi-query transformer models from multi-head checkpoints, 2023.
\newblock URL \url{https://arxiv.org/abs/2305.13245}.

\bibitem[Alvarez et~al.(2025)Alvarez, Almog, Chung, Layton, Stosic, Krashinsky, and Aubrey]{nvfp4}
Alvarez, E., Almog, O., Chung, E., Layton, S., Stosic, D., Krashinsky, R., and Aubrey, K.
\newblock Introducing nvfp4 for efficient and accurate low-precision inference, 2025.
\newblock URL \url{https://developer.nvidia.com/blog/introducing-nvfp4-for-efficient-and-accurate-low-precision-inference/}.

\bibitem[Ashkboos et~al.(2024)Ashkboos, Mohtashami, Croci, Li, Cameron, Jaggi, Alistarh, Hoefler, and Hensman]{quarot}
Ashkboos, S., Mohtashami, A., Croci, M.~L., Li, B., Cameron, P., Jaggi, M., Alistarh, D., Hoefler, T., and Hensman, J.
\newblock Quarot: Outlier-free 4-bit inference in rotated {LLM}s.
\newblock In \emph{The Thirty-eighth Annual Conference on Neural Information Processing Systems}, 2024.
\newblock URL \url{https://openreview.net/forum?id=dfqsW38v1X}.

\bibitem[Behnam et~al.(2025)Behnam, Fu, Zhao, Tsai, Yu, and Tumanov]{rocketkv}
Behnam, P., Fu, Y., Zhao, R., Tsai, P.-A., Yu, Z., and Tumanov, A.
\newblock Rocket{KV}: Accelerating long-context {LLM} inference via two-stage {KV} cache compression.
\newblock In \emph{Forty-second International Conference on Machine Learning}, 2025.
\newblock URL \url{https://openreview.net/forum?id=RyOpooIxDF}.

\bibitem[Chee et~al.(2023)Chee, Cai, Kuleshov, and De~Sa]{quip}
Chee, J., Cai, Y., Kuleshov, V., and De~Sa, C.~M.
\newblock Quip: 2-bit quantization of large language models with guarantees.
\newblock \emph{Advances in Neural Information Processing Systems}, 36:\penalty0 4396--4429, 2023.

\bibitem[Chmiel et~al.(2025)Chmiel, Fishman, Banner, and Soudry]{fp4train3}
Chmiel, B., Fishman, M., Banner, R., and Soudry, D.
\newblock Fp4 all the way: Fully quantized training of llms, 2025.
\newblock URL \url{https://arxiv.org/abs/2505.19115}.

\bibitem[Choromanski et~al.(2020)Choromanski, Likhosherstov, Dohan, Song, andreea, Sarlos, Hawkins, Davis, Mohiuddin, Kaiser, et~al.]{performers}
Choromanski, K., Likhosherstov, V., Dohan, D., Song, X., andreea, G., Sarlos, T., Hawkins, P., Davis, J., Mohiuddin, A., Kaiser, L., et~al.
\newblock Rethinking attention with performers.
\newblock \emph{arXiv preprint arXiv:2009.14794}, 2020.

\bibitem[Dao et~al.(2022)Dao, Fu, Ermon, Rudra, and R{\'e}]{fa1}
Dao, T., Fu, D., Ermon, S., Rudra, A., and R{\'e}, C.
\newblock Flashattention: Fast and memory-efficient exact attention with io-awareness.
\newblock \emph{Advances in neural information processing systems}, 35:\penalty0 16344--16359, 2022.

\bibitem[Darvish~Rouhani et~al.(2020)Darvish~Rouhani, Lo, Zhao, Liu, Fowers, Ovtcharov, Vinogradsky, Massengill, Yang, Bittner, et~al.]{msfp}
Darvish~Rouhani, B., Lo, D., Zhao, R., Liu, M., Fowers, J., Ovtcharov, K., Vinogradsky, A., Massengill, S., Yang, L., Bittner, R., et~al.
\newblock Pushing the limits of narrow precision inferencing at cloud scale with microsoft floating point.
\newblock \emph{Advances in neural information processing systems}, 33:\penalty0 10271--10281, 2020.

\bibitem[Darvish~Rouhani et~al.(2023)Darvish~Rouhani, Zhao, Elango, Shafipour, Hall, Mesmakhosroshahi, More, Melnick, Golub, Varatkar, et~al.]{shared_exponents}
Darvish~Rouhani, B., Zhao, R., Elango, V., Shafipour, R., Hall, M., Mesmakhosroshahi, M., More, A., Melnick, L., Golub, M., Varatkar, G., et~al.
\newblock With shared microexponents, a little shifting goes a long way.
\newblock In \emph{Proceedings of the 50th Annual International Symposium on Computer Architecture}, pp.\  1--13, 2023.

\bibitem[Dettmers et~al.(2022)Dettmers, Lewis, Belkada, and Zettlemoyer]{llmint8}
Dettmers, T., Lewis, M., Belkada, Y., and Zettlemoyer, L.
\newblock {LLM.int8()}: 8-bit matrix multiplication for transformers at scale.
\newblock NIPS '22, Red Hook, NY, USA, 2022. Curran Associates Inc.
\newblock ISBN 9781713871088.

\bibitem[Drumond et~al.(2018)Drumond, Lin, Jaggi, and Falsafi]{bfp1}
Drumond, M., Lin, T., Jaggi, M., and Falsafi, B.
\newblock Training dnns with hybrid block floating point.
\newblock \emph{Advances in Neural Information Processing Systems}, 31, 2018.

\bibitem[Dubey et~al.(2024)Dubey, Jauhri, Pandey, Kadian, Al-Dahle, Letman, Mathur, Schelten, Yang, Fan, et~al.]{llama3}
Dubey, A., Jauhri, A., Pandey, A., Kadian, A., Al-Dahle, A., Letman, A., Mathur, A., Schelten, A., Yang, A., Fan, A., et~al.
\newblock The llama 3 herd of models.
\newblock \emph{arXiv e-prints}, pp.\  arXiv--2407, 2024.

\bibitem[Frantar et~al.(2022)Frantar, Ashkboos, Hoefler, and Alistarh]{gptq}
Frantar, E., Ashkboos, S., Hoefler, T., and Alistarh, D.
\newblock Gptq: Accurate post-training quantization for generative pre-trained transformers.
\newblock \emph{arXiv preprint arXiv:2210.17323}, 2022.

\bibitem[Ge et~al.(2024)Ge, Zhang, Liu, Zhang, Han, and Gao]{discard}
Ge, S., Zhang, Y., Liu, L., Zhang, M., Han, J., and Gao, J.
\newblock Model tells you what to discard: Adaptive {KV} cache compression for {LLM}s.
\newblock In \emph{The Twelfth International Conference on Learning Representations}, 2024.
\newblock URL \url{https://openreview.net/forum?id=uNrFpDPMyo}.

\bibitem[Gholami et~al.(2022)Gholami, Kim, Dong, Yao, Mahoney, and Keutzer]{quant_survey}
Gholami, A., Kim, S., Dong, Z., Yao, Z., Mahoney, M.~W., and Keutzer, K.
\newblock A survey of quantization methods for efficient neural network inference.
\newblock In \emph{Low-power computer vision}, pp.\  291--326. Chapman and Hall/CRC, 2022.

\bibitem[Guo et~al.(2025)Guo, Yang, Zhang, Song, Zhang, Xu, Zhu, Ma, Wang, Bi, et~al.]{guo2025deepseek}
Guo, D., Yang, D., Zhang, H., Song, J., Zhang, R., Xu, R., Zhu, Q., Ma, S., Wang, P., Bi, X., et~al.
\newblock Deepseek-r1: Incentivizing reasoning capability in llms via reinforcement learning.
\newblock \emph{arXiv preprint arXiv:2501.12948}, 2025.

\bibitem[He et~al.(2024)He, Zhang, Wu, Liu, Zhou, and Zhuang]{zipcache}
He, Y., Zhang, L., Wu, W., Liu, J., Zhou, H., and Zhuang, B.
\newblock Zipcache: Accurate and efficient kv cache quantization with salient token identification.
\newblock \emph{Advances in Neural Information Processing Systems}, 37:\penalty0 68287--68307, 2024.

\bibitem[Hendrycks et~al.(2020)Hendrycks, Burns, Basart, Zou, Mazeika, Song, and Steinhardt]{mmlu}
Hendrycks, D., Burns, C., Basart, S., Zou, A., Mazeika, M., Song, D., and Steinhardt, J.
\newblock Measuring massive multitask language understanding.
\newblock \emph{arXiv preprint arXiv:2009.03300}, 2020.

\bibitem[Hendrycks et~al.(2021)Hendrycks, Burns, Kadavath, Arora, Basart, Tang, Song, and Steinhardt]{math500}
Hendrycks, D., Burns, C., Kadavath, S., Arora, A., Basart, S., Tang, E., Song, D., and Steinhardt, J.
\newblock Measuring mathematical problem solving with the math dataset.
\newblock \emph{arXiv preprint arXiv:2103.03874}, 2021.

\bibitem[Hooper et~al.(2024)Hooper, Kim, Mohammadzadeh, Mahoney, Shao, Keutzer, and Gholami]{kvquant}
Hooper, C. R.~C., Kim, S., Mohammadzadeh, H., Mahoney, M.~W., Shao, S., Keutzer, K., and Gholami, A.
\newblock {KVQ}uant: Towards 10 million context length {LLM} inference with {KV} cache quantization.
\newblock In \emph{The Thirty-eighth Annual Conference on Neural Information Processing Systems}, 2024.
\newblock URL \url{https://openreview.net/forum?id=0LXotew9Du}.

\bibitem[Jacob et~al.(2018{\natexlab{a}})Jacob, Kligys, Chen, Zhu, Tang, andrew, Adam, and Kalenichenko]{fixed_point2}
Jacob, B., Kligys, S., Chen, B., Zhu, M., Tang, M., andrew, H., Adam, H., and Kalenichenko, D.
\newblock Quantization and training of neural networks for efficient integer-arithmetic-only inference.
\newblock In \emph{Proceedings of the IEEE conference on computer vision and pattern recognition}, pp.\  2704--2713, 2018{\natexlab{a}}.

\bibitem[Jacob et~al.(2018{\natexlab{b}})Jacob, Kligys, Chen, Zhu, Tang, andrew, Adam, and Kalenichenko]{qat2}
Jacob, B., Kligys, S., Chen, B., Zhu, M., Tang, M., andrew, H., Adam, H., and Kalenichenko, D.
\newblock Quantization and training of neural networks for efficient integer-arithmetic-only inference.
\newblock In \emph{2018 IEEE/CVF Conference on Computer Vision and Pattern Recognition}, pp.\  2704--2713, 2018{\natexlab{b}}.
\newblock \doi{10.1109/CVPR.2018.00286}.

\bibitem[Jia(2024)]{aime_2024}
Jia, M.
\newblock Aime problem set 2024, 2024.
\newblock URL \url{https://huggingface.co/datasets/Maxwell-Jia/AIME_2024}.

\bibitem[Jiang et~al.(2024)Jiang, Li, Zhang, Wu, Luo, Ahn, Han, Abdi, Li, Lin, et~al.]{spa1}
Jiang, H., Li, Y., Zhang, C., Wu, Q., Luo, X., Ahn, S., Han, Z., Abdi, A.~H., Li, D., Lin, C.-Y., et~al.
\newblock Minference 1.0: Accelerating pre-filling for long-context llms via dynamic sparse attention.
\newblock \emph{Advances in Neural Information Processing Systems}, 37:\penalty0 52481--52515, 2024.

\bibitem[Kang et~al.(2024)Kang, Zhang, Kundu, Jeong, Liu, Krishna, and Zhao]{gear}
Kang, H., Zhang, Q., Kundu, S., Jeong, G., Liu, Z., Krishna, T., and Zhao, T.
\newblock Gear: An efficient kv cache compression recipe for near-lossless generative inference of llm, 2024.
\newblock URL \url{https://arxiv.org/abs/2403.05527}.

\bibitem[Kwon et~al.(2023)Kwon, Li, Zhuang, Sheng, Zheng, Yu, Gonzalez, Zhang, and Stoica]{kwon2023vllm}
Kwon, W., Li, Z., Zhuang, S., Sheng, Y., Zheng, L., Yu, C.~H., Gonzalez, J., Zhang, H., and Stoica, I.
\newblock Efficient memory management for large language model serving with pagedattention.
\newblock In \emph{Proceedings of the 29th symposium on operating systems principles}, pp.\  611--626, 2023.

\bibitem[Lefaudeux et~al.(2022)Lefaudeux, Massa, Liskovich, Xiong, Caggiano, Naren, Xu, Hu, Tintore, Zhang, Labatut, Haziza, Wehrstedt, Reizenstein, and Sizov]{xFormers2022}
Lefaudeux, B., Massa, F., Liskovich, D., Xiong, W., Caggiano, V., Naren, S., Xu, M., Hu, J., Tintore, M., Zhang, S., Labatut, P., Haziza, D., Wehrstedt, L., Reizenstein, J., and Sizov, G.
\newblock xformers: A modular and hackable transformer modelling library.
\newblock \url{https://github.com/facebookresearch/xformers}, 2022.

\bibitem[Li et~al.(2024{\natexlab{a}})Li, Lin, Zhang, Cai, Li, Guo, Xie, Meng, Zhu, and Han]{svdquant}
Li, M., Lin, Y., Zhang, Z., Cai, T., Li, X., Guo, J., Xie, E., Meng, C., Zhu, J.-Y., and Han, S.
\newblock Svdquant: Absorbing outliers by low-rank components for 4-bit diffusion models.
\newblock \emph{arXiv preprint arXiv:2411.05007}, 2024{\natexlab{a}}.

\bibitem[Li et~al.(2025)Li, XING, Li, Qu, Zhen, Yao, Liu, Pan, and Yuan]{kvtuner}
Li, X., XING, Z., Li, Y., Qu, L., Zhen, H.-L., Yao, Y., Liu, W., Pan, S.~J., and Yuan, M.
\newblock {KVT}uner: Sensitivity-aware layer-wise mixed-precision {KV} cache quantization for efficient and nearly lossless {LLM} inference.
\newblock In \emph{Forty-second International Conference on Machine Learning}, 2025.
\newblock URL \url{https://openreview.net/forum?id=zDwipF6h06}.

\bibitem[Li et~al.(2024{\natexlab{b}})Li, Huang, Yang, Venkitesh, Locatelli, Ye, Cai, Lewis, and Chen]{snapkv}
Li, Y., Huang, Y., Yang, B., Venkitesh, B., Locatelli, A., Ye, H., Cai, T., Lewis, P., and Chen, D.
\newblock Snap{KV}: {LLM} knows what you are looking for before generation.
\newblock In \emph{The Thirty-eighth Annual Conference on Neural Information Processing Systems}, 2024{\natexlab{b}}.
\newblock URL \url{https://openreview.net/forum?id=poE54GOq2l}.

\bibitem[Lin et~al.(2016)Lin, Talathi, and Annapureddy]{fixed_point_1}
Lin, D., Talathi, S., and Annapureddy, S.
\newblock Fixed point quantization of deep convolutional networks.
\newblock In Balcan, M.~F. and Weinberger, K.~Q. (eds.), \emph{Proceedings of The 33rd International Conference on Machine Learning}, volume~48 of \emph{Proceedings of Machine Learning Research}, pp.\  2849--2858, New York, New York, USA, 20--22 Jun 2016. PMLR.
\newblock URL \url{https://proceedings.mlr.press/v48/linb16.html}.

\bibitem[Lin et~al.(2024)Lin, Tang, Tang, Yang, Chen, Wang, Xiao, Dang, Gan, and Han]{awq}
Lin, J., Tang, J., Tang, H., Yang, S., Chen, W.-M., Wang, W.-C., Xiao, G., Dang, X., Gan, C., and Han, S.
\newblock Awq: Activation-aware weight quantization for on-device llm compression and acceleration.
\newblock In Gibbons, P., Pekhimenko, G., and Sa, C.~D. (eds.), \emph{Proceedings of Machine Learning and Systems}, volume~6, pp.\  87--100, 2024.
\newblock URL \url{https://proceedings.mlsys.org/paper_files/paper/2024/file/42a452cbafa9dd64e9ba4aa95cc1ef21-Paper-Conference.pdf}.

\bibitem[Lin* et~al.(2024)Lin*, Tang*, Yang*, Zhang, Xiao, Gan, and Han]{qserve}
Lin*, Y., Tang*, H., Yang*, S., Zhang, Z., Xiao, G., Gan, C., and Han, S.
\newblock Qserve: W4a8kv4 quantization and system co-design for efficient llm serving.
\newblock \emph{arXiv preprint arXiv:2405.04532}, 2024.

\bibitem[Liu et~al.(2024{\natexlab{a}})Liu, Feng, Wang, Wang, Liu, Zhao, Dengr, Ruan, Dai, Guo, et~al.]{mhla}
Liu, A., Feng, B., Wang, B., Wang, B., Liu, B., Zhao, C., Dengr, C., Ruan, C., Dai, D., Guo, D., et~al.
\newblock Deepseek-v2: A strong, economical and efficient mixture-of-experts language model.
\newblock \emph{arXiv preprint arXiv:2405.04434}, 2024{\natexlab{a}}.

\bibitem[Liu et~al.(2024{\natexlab{b}})Liu, Liu, Pan, He, Haffari, and Zhuang]{minicache}
Liu, A., Liu, J., Pan, Z., He, Y., Haffari, G., and Zhuang, B.
\newblock Minicache: Kv cache compression in depth dimension for large language models.
\newblock \emph{Advances in Neural Information Processing Systems}, 37:\penalty0 139997--140031, 2024{\natexlab{b}}.

\bibitem[Liu et~al.(2024{\natexlab{c}})Liu, Bai, Lin, Li, Gao, Xu, Hou, Yao, and Yuan]{intactkv}
Liu, R., Bai, H., Lin, H., Li, Y., Gao, H., Xu, Z., Hou, L., Yao, J., and Yuan, C.
\newblock {I}ntact{KV}: Improving large language model quantization by keeping pivot tokens intact.
\newblock In Ku, L.-W., andre, M., and Srikumar, V. (eds.), \emph{Findings of the Association for Computational Linguistics: ACL 2024}, pp.\  7716--7741, Bangkok, Thailand, August 2024{\natexlab{c}}. Association for Computational Linguistics.
\newblock \doi{10.18653/v1/2024.findings-acl.460}.
\newblock URL \url{https://aclanthology.org/2024.findings-acl.460/}.

\bibitem[Liu et~al.(2025)Liu, Sun, Zhang, Bai, Yu, Yu, Yuan, and Hou]{reasoninghurtsquant}
Liu, R., Sun, Y., Zhang, M., Bai, H., Yu, X., Yu, T., Yuan, C., and Hou, L.
\newblock Quantization hurts reasoning? an empirical study on quantized reasoning models, 2025.
\newblock URL \url{https://arxiv.org/abs/2504.04823}.

\bibitem[Liu et~al.(2024{\natexlab{d}})Liu, Yuan, Jin, Zhong, Xu, Braverman, Chen, and Hu]{kivi}
Liu, Z., Yuan, J., Jin, H., Zhong, S.~H., Xu, Z., Braverman, V., Chen, B., and Hu, X.
\newblock Kivi: a tuning-free asymmetric 2bit quantization for kv cache.
\newblock In \emph{Proceedings of the 41st International Conference on Machine Learning}, ICML'24. JMLR.org, 2024{\natexlab{d}}.

\bibitem[Meta()]{PyTorch}
Meta.
\newblock Pytorch.
\newblock URL \url{https://pytorch.org/}.

\bibitem[Micikevicius et~al.(2022)Micikevicius, Stosic, Burgess, Cornea, Dubey, Grisenthwaite, Ha, Heinecke, Judd, Kamalu, et~al.]{fp8_nvidia}
Micikevicius, P., Stosic, D., Burgess, N., Cornea, M., Dubey, P., Grisenthwaite, R., Ha, S., Heinecke, A., Judd, P., Kamalu, J., et~al.
\newblock Fp8 formats for deep learning.
\newblock \emph{arXiv preprint arXiv:2209.05433}, 2022.

\bibitem[Mu et~al.(2023)Mu, Li, and Goodman]{gist}
Mu, J., Li, X., and Goodman, N.
\newblock Learning to compress prompts with gist tokens.
\newblock \emph{Advances in Neural Information Processing Systems}, 36:\penalty0 19327--19352, 2023.

\bibitem[NVIDIA()]{tensorrt}
NVIDIA.
\newblock Working with quantized types.
\newblock URL \url{https://docs.nvidia.com/deeplearning/tensorrt/latest/inference-library/work-quantized-types.html}.

\bibitem[NVIDIA(2025{\natexlab{a}})]{blackwell_infer}
NVIDIA.
\newblock Nvidia blackwell delivers world-record deepseek-r1 inference performance, 2025{\natexlab{a}}.
\newblock URL \url{https://developer.nvidia.com/blog/nvidia-blackwell-delivers-world-record-deepseek-r1-inference-performance/}.

\bibitem[NVIDIA(2025{\natexlab{b}})]{scale_dims}
NVIDIA.
\newblock Ptx warp-level block scaling, 2025{\natexlab{b}}.
\newblock URL \url{https://docs.nvidia.com/cuda/parallel-thread-execution/#warp-level-block-scaling}.

\bibitem[NVIDIA(2025{\natexlab{c}})]{trt_opt}
NVIDIA.
\newblock {TensorRT-Model-Optimizer}, 2025{\natexlab{c}}.
\newblock URL \url{https://github.com/NVIDIA/TensorRT-Model-Optimizer/tree/main}.

\bibitem[Rein et~al.(2023)Rein, Hou, Stickland, Petty, Pang, Dirani, Michael, and Bowman]{gpqa}
Rein, D., Hou, B.~L., Stickland, A.~C., Petty, J., Pang, R.~Y., Dirani, J., Michael, J., and Bowman, S.~R.
\newblock Gpqa: A graduate-level google-proof q\&a benchmark, 2023.

\bibitem[Rouhani et~al.(2025)Rouhani, Garegrat, Savell, More, Han, Zhao, Hall, Klar, Chung, Yu, Schulte, Wittig, Bratt, Stephens, Milanovic, Brothers, Dubey, Cornea, Heinecke, andres, Langhammer, Deng, Naumov, Micikevicius, Siu, and Verrilli]{mxfp4}
Rouhani, B.~D., Garegrat, N., Savell, T., More, A., Han, K.-N., Zhao, R., Hall, M., Klar, J., Chung, E., Yu, Y., Schulte, M., Wittig, R., Bratt, I., Stephens, N., Milanovic, J., Brothers, J., Dubey, P., Cornea, M., Heinecke, A., andres, R., Langhammer, M., Deng, S., Naumov, M., Micikevicius, P., Siu, M., and Verrilli, C.
\newblock Ocp microscaling formats (mx) specification, 2025.
\newblock URL \url{https://www.opencompute.org/documents/ocp-microscaling-formats-mx-v1-0-spec-final-pdf}.

\bibitem[Salesforce()]{Wikitext-2}
Salesforce.
\newblock Wikitext-2 dataset.
\newblock URL \url{https://huggingface.co/datasets/Salesforce/wikitext/viewer/wikitext-2-v1}.

\bibitem[Shah et~al.(2024)Shah, Bikshandi, Zhang, Thakkar, Ramani, and Dao]{fa3}
Shah, J., Bikshandi, G., Zhang, Y., Thakkar, V., Ramani, P., and Dao, T.
\newblock Flashattention-3: Fast and accurate attention with asynchrony and low-precision.
\newblock \emph{Advances in Neural Information Processing Systems}, 37:\penalty0 68658--68685, 2024.

\bibitem[Shazeer(2019)]{mqa}
Shazeer, N.
\newblock Fast transformer decoding: One write-head is all you need.
\newblock \emph{arXiv preprint arXiv:1911.02150}, 2019.

\bibitem[Stock et~al.(2021)Stock, Fan, Graham, Grave, Gribonval, Jegou, and Joulin]{qat1}
Stock, P., Fan, A., Graham, B., Grave, E., Gribonval, R., Jegou, H., and Joulin, A.
\newblock Training with quantization noise for extreme model compression.
\newblock In \emph{International Conference on Learning Representations}, 2021.
\newblock URL \url{https://openreview.net/forum?id=dV19Yyi1fS3}.

\bibitem[Team(2024)]{genmo2024mochi}
Team, G.
\newblock Mochi 1.
\newblock \url{https://github.com/genmoai/models}, 2024.

\bibitem[Tseng et~al.(2024)Tseng, Chee, Sun, Kuleshov, and De~Sa]{quipsharp}
Tseng, A., Chee, J., Sun, Q., Kuleshov, V., and De~Sa, C.
\newblock Quip\#: Even better llm quantization with hadamard incoherence and lattice codebooks.
\newblock \emph{arXiv preprint arXiv:2402.04396}, 2024.

\bibitem[Tseng et~al.(2025)Tseng, Yu, and Park]{fp4train2}
Tseng, A., Yu, T., and Park, Y.
\newblock Training llms with mxfp4, 2025.
\newblock URL \url{https://arxiv.org/abs/2502.20586}.

\bibitem[vLLM()]{vLLM}
vLLM.
\newblock Quantization.
\newblock URL \url{https://docs.vllm.ai/en/latest/features/quantization/index.html}.

\bibitem[Wang et~al.(2025)Wang, Gong, Liu, Zhao, Yang, Guo, Zha, and CHENG]{fp4train1}
Wang, R., Gong, Y., Liu, X., Zhao, G., Yang, Z., Guo, B., Zha, Z.-J., and CHENG, P.
\newblock Optimizing large language model training using {FP}4 quantization.
\newblock In \emph{Forty-second International Conference on Machine Learning}, 2025.
\newblock URL \url{https://openreview.net/forum?id=uK7JArZEJM}.

\bibitem[Xiao et~al.(2023{\natexlab{a}})Xiao, Lin, Seznec, Wu, Demouth, and Han]{smoothquant}
Xiao, G., Lin, J., Seznec, M., Wu, H., Demouth, J., and Han, S.
\newblock Smoothquant: Accurate and efficient post-training quantization for large language models.
\newblock In \emph{International conference on machine learning}, pp.\  38087--38099. PMLR, 2023{\natexlab{a}}.

\bibitem[Xiao et~al.(2023{\natexlab{b}})Xiao, Tian, Chen, Han, and Lewis]{spa5}
Xiao, G., Tian, Y., Chen, B., Han, S., and Lewis, M.
\newblock Efficient streaming language models with attention sinks.
\newblock \emph{arXiv preprint arXiv:2309.17453}, 2023{\natexlab{b}}.

\bibitem[Yang et~al.(2025)Yang, Li, Yang, Zhang, Hui, Zheng, Yu, Gao, Huang, Lv, Zheng, Liu, Zhou, Huang, Hu, Ge, Wei, Lin, Tang, Yang, Tu, Zhang, Yang, Yang, Zhou, Zhou, Lin, Dang, Bao, Yang, Yu, Deng, Li, Xue, Li, Zhang, Wang, Zhu, Men, Gao, Liu, Luo, Li, Tang, Yin, Ren, Wang, Zhang, Ren, Fan, Su, Zhang, Zhang, Wan, Liu, Wang, Cui, Zhang, Zhou, and Qiu]{qwen3}
Yang, A., Li, A., Yang, B., Zhang, B., Hui, B., Zheng, B., Yu, B., Gao, C., Huang, C., Lv, C., Zheng, C., Liu, D., Zhou, F., Huang, F., Hu, F., Ge, H., Wei, H., Lin, H., Tang, J., Yang, J., Tu, J., Zhang, J., Yang, J., Yang, J., Zhou, J., Zhou, J., Lin, J., Dang, K., Bao, K., Yang, K., Yu, L., Deng, L., Li, M., Xue, M., Li, M., Zhang, P., Wang, P., Zhu, Q., Men, R., Gao, R., Liu, S., Luo, S., Li, T., Tang, T., Yin, W., Ren, X., Wang, X., Zhang, X., Ren, X., Fan, Y., Su, Y., Zhang, Y., Zhang, Y., Wan, Y., Liu, Y., Wang, Z., Cui, Z., Zhang, Z., Zhou, Z., and Qiu, Z.
\newblock Qwen3 technical report, 2025.
\newblock URL \url{https://arxiv.org/abs/2505.09388}.

\bibitem[Yang et~al.(2024{\natexlab{a}})Yang, Kim, Bae, Kwon, Park, Yang, Kwon, and Lee]{mikv}
Yang, J.~Y., Kim, B., Bae, J., Kwon, B., Park, G., Yang, E., Kwon, S.~J., and Lee, D.
\newblock No token left behind: Reliable kv cache compression via importance-aware mixed precision quantization, 2024{\natexlab{a}}.
\newblock URL \url{https://arxiv.org/abs/2402.18096}.

\bibitem[Yang et~al.(2024{\natexlab{b}})Yang, Cao, Chen, Qin, Yang, Zhao, and Chen]{kvsharer}
Yang, Y., Cao, Z., Chen, Q., Qin, L., Yang, D., Zhao, H., and Chen, Z.
\newblock Kvsharer: Efficient inference via layer-wise dissimilar kv cache sharing.
\newblock \emph{arXiv preprint arXiv:2410.18517}, 2024{\natexlab{b}}.

\bibitem[Yang et~al.(2024{\natexlab{c}})Yang, Bao, Kong, Wu, He, Zhai, Wu, Zhao, Yu, Lv, Mao, Huang, Feng, Zhang, Chen, Chen, Ding, Wang, Hong, Chen, Shang, Lu, Tang, Chen, and Wang]{yang2024cogvideox}
Yang, Z., Bao, F., Kong, Z., Wu, Z., He, J., Zhai, H., Wu, J., Zhao, X., Yu, H., Lv, H., Mao, Q., Huang, Y., Feng, X., Zhang, C., Chen, J., Chen, W., Ding, W., Wang, C., Hong, W., Chen, W., Shang, S., Lu, X., Tang, J., Chen, W., and Wang, Y.
\newblock Cogvideox: Text-to-video diffusion models with an expert transformer.
\newblock \emph{arXiv preprint arXiv:2408.06072}, 2024{\natexlab{c}}.

\bibitem[Yao et~al.(2022)Yao, Yazdani~Aminabadi, Zhang, Wu, Li, and He]{zeroquant}
Yao, Z., Yazdani~Aminabadi, R., Zhang, M., Wu, X., Li, C., and He, Y.
\newblock Zeroquant: Efficient and affordable post-training quantization for large-scale transformers.
\newblock \emph{Advances in neural information processing systems}, 35:\penalty0 27168--27183, 2022.

\bibitem[Yue et~al.(2024)Yue, Yuan, Duanmu, Zhou, Wu, and Nie]{wkvquant}
Yue, Y., Yuan, Z., Duanmu, H., Zhou, S., Wu, J., and Nie, L.
\newblock Wkvquant: Quantizing weight and key/value cache for large language models gains more.
\newblock \emph{arXiv preprint arXiv:2402.12065}, 2024.

\bibitem[Zhang et~al.()Zhang, Huang, Zhang, Zhu, Chen, et~al.]{sageattention2}
Zhang, J., Huang, H., Zhang, P., Zhu, J., Chen, J., et~al.
\newblock Sageattention2: Efficient attention with smoothing q and per-thread quantization.
\newblock In \emph{First Workshop on Scalable Optimization for Efficient and Adaptive Foundation Models}.

\bibitem[Zhang et~al.(2024)Zhang, Wei, Huang, Zhang, Zhu, and Chen]{sageattention1}
Zhang, J., Wei, J., Huang, H., Zhang, P., Zhu, J., and Chen, J.
\newblock Sageattention: Accurate 8-bit attention for plug-and-play inference acceleration.
\newblock \emph{arXiv preprint arXiv:2410.02367}, 2024.

\bibitem[Zhang et~al.(2025{\natexlab{a}})Zhang, Wei, Zhang, Xu, Huang, Wang, Jiang, Zhu, and Chen]{sageattention3}
Zhang, J., Wei, J., Zhang, P., Xu, X., Huang, H., Wang, H., Jiang, K., Zhu, J., and Chen, J.
\newblock Sageattention3: Microscaling fp4 attention for inference and an exploration of 8-bit training.
\newblock \emph{arXiv preprint arXiv:2505.11594}, 2025{\natexlab{a}}.

\bibitem[Zhang et~al.(2025{\natexlab{b}})Zhang, Xiang, Huang, Wei, Xi, Zhu, and Chen]{spa2}
Zhang, J., Xiang, C., Huang, H., Wei, J., Xi, H., Zhu, J., and Chen, J.
\newblock Spargeattn: Accurate sparse attention accelerating any model inference.
\newblock \emph{arXiv preprint arXiv:2502.18137}, 2025{\natexlab{b}}.

\bibitem[Zhang et~al.(2022)Zhang, McDanel, and Kung]{fast}
Zhang, S.~Q., McDanel, B., and Kung, H.
\newblock Fast: Dnn training under variable precision block floating point with stochastic rounding.
\newblock In \emph{2022 IEEE International Symposium on High-Performance Computer Architecture (HPCA)}, pp.\  846--860. IEEE, 2022.

\bibitem[Zhang et~al.(2023)Zhang, Sheng, Zhou, Chen, Zheng, Cai, Song, Tian, R{\'e}, Barrett, et~al.]{h20}
Zhang, Z., Sheng, Y., Zhou, T., Chen, T., Zheng, L., Cai, R., Song, Z., Tian, Y., R{\'e}, C., Barrett, C., et~al.
\newblock H2o: Heavy-hitter oracle for efficient generative inference of large language models.
\newblock \emph{Advances in Neural Information Processing Systems}, 36:\penalty0 34661--34710, 2023.

\end{thebibliography}
\bibliographystyle{mlsys2025}

\newpage
\appendix
\newpage
\onecolumn
\section{Proof of magnitude reduction transformation}
\label{appendix:algo}

Let the original query and key matrices be $Q, K \in \mathbb{R}^{B \times d}$. 
We introduce a pair of linear transformations:
\[
Q' = Q R^{-T}, \qquad K' = K R,
\]
where $R \in \mathbb{R}^{d \times d}$ is an invertible transformation. 
This transformation preserves the attention scores:
\[
Q' {K'}^\top = (Q R^{-T})(K R)^\top = Q R^{-T} R^\top K^\top = Q K^\top.
\]

Let the second-moment (or Hessian) matrices of $Q$ and $K$ be:
\[
X = \mathbb{E}[Q^\top Q], \qquad Y = \mathbb{E}[K^\top K],
\]
which capture the energy distributions of $Q$ and $K$ along each feature direction.

After applying the reparameterization $R$, the average squared magnitudes of the transformed queries and keys become:
\[
\mathbb{E}[\|Q'\|^2] = \mathrm{tr}(R^{-1} X R^{-T}), \qquad 
\mathbb{E}[\|K'\|^2] = \mathrm{tr}(R^\top Y R).
\]
To jointly minimize their overall energy (and hence the quantization error), 
we minimize the product of these quantities:
\begin{equation}
\label{eq:joint-objective}
\min_R \; \mathrm{tr}(R^{-1} X R^{-T}) \cdot \mathrm{tr}(R^\top Y R).
\end{equation}

We solve~\eqref{eq:joint-objective} analytically using singular value decomposition (SVD).
Let
\[
X^{1/2} Y^{1/2} = U S V^\top
\]
be the SVD of the symmetric positive semi-definite matrix product $X^{1/2} Y^{1/2}$,
where $U, V$ are orthogonal and $S = \mathrm{diag}(s_i)$ contains the singular values.

Define the optimal transformation as:
\[
R = Y^{-1/2} V S^{1/2}, \qquad R^{-1} = S^{-1/2} V^\top Y^{1/2}.
\]
such that it makes the traces equal.
\begin{align*}
\mathrm{tr}(R^{-1} X R^{-T}) 
&= \mathrm{tr}\!\left(S^{-1/2} V^\top Y^{1/2} X Y^{1/2} V S^{-1/2}\right)
\\&= \mathrm{tr}(S), \\[4pt]
\mathrm{tr}(R^\top Y R)
&= \mathrm{tr}\!\left(S^{1/2} V^\top Y^{-1/2} Y Y^{-1/2} V S^{1/2}\right)
\\&= \mathrm{tr}(S).
\end{align*}
Hence, the minimized joint energy becomes:
\[
\mathrm{tr}(R^{-1} X R^{-T}) \cdot \mathrm{tr}(R^\top Y R) = \left(\mathrm{tr}(S)\right)^2.
\]
This is the minimum achievable value of~\eqref{eq:joint-objective}, 
and the corresponding $R$ aligns the spectral bases of $X$ and $Y$ optimally.



\end{document}